\documentclass[10pt,twocolumn,letterpaper]{article}

\usepackage[pagenumbers]{cvpr} 

\definecolor{cvprblue}{rgb}{0.21,0.49,0.74}
\usepackage[pagebackref,breaklinks,colorlinks,allcolors=cvprblue]{hyperref}

\usepackage[accsupp]{axessibility}  
\usepackage{graphicx}
\usepackage{wrapfig}
\usepackage{booktabs}
\usepackage{multirow}
\usepackage{caption}
\usepackage{xcolor}         
\newcommand{\best}[1]{\textbf{#1}}

\def\paperID{22581} 
\def\confName{CVPR}
\def\confYear{2026}

\title{iLRM: An Iterative Large 3D Reconstruction Model}

\author{
Gyeongjin Kang$^{1}$\quad
Seungtae Nam$^{2}$\quad
Seungkwon Yang$^{2}$\quad
Xiangyu Sun$^{1}$\\
Sameh Khamis$^{3}$\quad
Abdelrahman Mohamed$^{4}$\footnotemark\quad
Eunbyung Park$^{2}$
\vspace{2mm} \\
$^1$Sungkyunkwan University\hspace{2.4mm}$^2$Yonsei University\hspace{2.4mm}$^3$Rembrand\hspace{2.4mm}$^4$Meta
\vspace{2mm} \\
{\small \url{https://gynjn.github.io/iLRM/}}
}

\begin{document}
\twocolumn[{%
\renewcommand\twocolumn[1][]{#1}%
\maketitle
\begin{center}
    \centering
    \captionsetup{type=figure}
    \vspace{-1.0em}
    \includegraphics[width=1.0\linewidth]{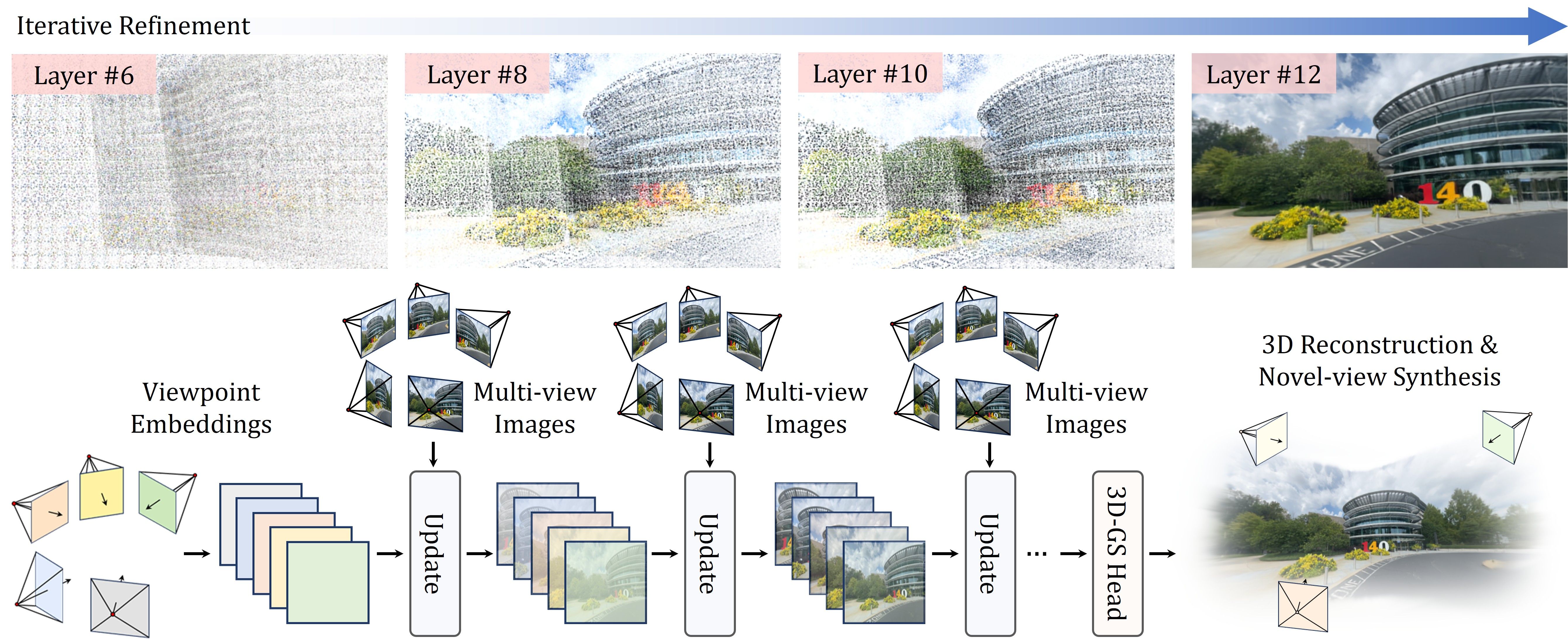}
    \vspace*{-4mm}
    \caption{The overall architecture of the \textit{iLRM}. As the layer index increases, compact viewpoint tokens are iteratively refined by attending to multi-view image tokens and are finally decoded into 3D Gaussian primitives, enabling efficient and high quality 3D reconstruction.}
    \label{fig:teaser}
\end{center}%
}]

\renewcommand{\thefootnote}{*}
\footnotetext{Work done while at Rembrand.}
\begin{abstract}
Feed-forward 3D modeling has emerged as a promising approach for rapid and high-quality 3D reconstruction. In particular, directly generating explicit 3D representations, such as 3D Gaussian splatting, has attracted significant attention due to its fast and high-quality rendering. However, many state-of-the-art methods, primarily based on transformer architectures, suffer from severe scalability issues because they rely on full attention across image tokens from multiple input views, resulting in prohibitive computational costs as the number of views or image resolution increases. Toward a scalable and efficient feed-forward 3D reconstruction, we introduce an iterative Large 3D Reconstruction Model (\textit{iLRM}) that generates 3D Gaussian representations through an iterative refinement mechanism, guided by three core principles: (1) decoupling the scene representation from input images to enable \textit{compact 3D representations}; (2) decomposing global multi-view interactions into a \textit{two-stage attention} scheme to reduce computational costs; and (3) injecting \textit{high-resolution information at every layer} to achieve high-fidelity reconstruction.
Experimental results on widely used datasets, such as RE10K and DL3DV, demonstrate that \textit{iLRM} outperforms existing methods in both reconstruction quality and speed.
\end{abstract}

\section{Introduction}
\vspace{-1mm}
Since the recent success of 3D Gaussian Splatting (3D-GS)~\cite{kerbl20233dgs}, significant progress has been made in leveraging this 3D representation for building generalizable feed-forward 3D reconstruction models~\cite{charatan2024pixelsplat, tang2025lgm, xu2024grm, chen2025mvsplat, chen2024mvsplat360, xu2025depthsplat, zhang2025gs-lrm}.
These methods typically train large neural networks to transform multi-view input images into feature representations, then regress Gaussian attributes. 
In contrast to per-scene 3D-GS optimization approaches~\cite{kerbl20233dgs, mallick2024taming, lu2024turbo-gs, fang2024mini-splat}, these feed-forward models can reconstruct 3D scenes in a single forward pass, offering near real-time performance.
Moreover, the prior knowledge learned from large-scale datasets~\cite{zhou2018re10k, ling2024dl3dv, deitke2023objaverse, deitke2024objaverse-xl} allows them to effectively generalize to unseen scenes. 
While their reconstruction quality often lags behind that of per-scene optimization methods, fast reconstruction speed and generalization capability mark a promising step toward the long-standing goal of achieving accurate and real-time 3D scene reconstruction.

Among the promising approaches, pixel-aligned Gaussian models~\cite{charatan2024pixelsplat, szymanowicz2024splatter, zheng2024gps} have emerged as the de facto standard, leveraging decades of advances 
developed for numerous image-based tasks. While these models have proven effective, they also exhibit certain limitations. In particular, since they generate per-pixel Gaussians directly from the input images, the image resolution determines the number of Gaussians produced, which can lead to an excessive number of redundant Gaussians. For example, given input images at 1K resolution across 200 viewpoints (a scale comparable to the bicycle scene in the mip-NeRF 360 dataset~\cite{barron2022mip}), these methods would produce 200 million Gaussians, despite prior studies~\cite{lee2024compact, fan2024lightgaussian, chen2024hac, lee2025optimized} demonstrating that such scenes can be efficiently represented with around 0.5 million Gaussians. To mitigate this issue, several techniques have been proposed, such as Gaussian regularization~\cite{ziwen2025llrm} and feature fusion~\cite{wang2025freesplat}. Alternatively, the network 
can also be designed to generate fewer Gaussians, for example by downsampling the output resolutions. However, these strategies still require processing high-resolution 
images and therefore do not address another fundamental limitation of these models: high computational and memory demands.

A significant portion of computational and memory overhead arises from modeling interactions across multiple input views in feed-forward 3D reconstruction models. For instance, GS-LRM~\cite{zhang2025gs-lrm} performs full attention over all image tokens from every input view, leading to a quadratic increase in complexity with respect to both the number of views and image resolution. MVSplat~\cite{chen2025mvsplat} and DepthSplat~\cite{xu2025depthsplat} construct and process cost volumes for each view, further contributing to the computational demands. While one might attempt to alleviate this burden by reducing the input image resolution or using a sparser set of views, such strategies risk discarding essential geometric and appearance information required for accurate reconstruction. 

Beyond the computational complexity and the inefficiency of the generated representations, we also question whether the prevailing formulation, casting 3D reconstruction as a sequence-to-sequence problem that maps entire sets of image tokens to dense, pixel-aligned Gaussians, is fundamentally well-suited to the nature of the task.
While this formulation has achieved impressive results~\cite{zhang2025gs-lrm, ziwen2025llrm, jin2024lvsm}, it remains primarily a one-shot 3D scene generation process. In contrast, the recent optimization-based methods~\cite{kerbl20233dgs, mallick2024taming} follow a fundamentally different strategy: they treat reconstruction as an iterative refinement process, where each iteration involves rendering the current scene estimate, measuring reconstruction error, and updating the representation accordingly. This loop enables the model to progressively capture finer geometric and appearance details while ensuring strong 3D consistency. The success of these methods suggests that high-quality 3D reconstruction may benefit not only from expressive representations but also from feedback-driven iterative refinement, a trait largely absent in existing feed-forward 3D models.

In this paper, we introduce \textit{iLRM}\footnote[1]{By “feedback-driven,” we mean that at every update layer the current scene tokens are explicitly revised via cross-attention with each image tokens. We call this “iterative” because the scene representation is repeatedly updated layer by layer under static (unchanged) image evidence, rather than merely transformed by stacked self-attention blocks~\cite{jaegle2021perceiver}.}, an iterative large 3D reconstruction model that effectively 1) incorporates the principles of feedback-driven refinement, while also 2) addressing the computational burden and representational inefficiencies inherent in existing feed-forward approaches.
As illustrated in Fig.~\ref{fig:teaser}, the network (acting as an optimizer) transforms the embedding features (analogous to updating the 3D-GS representation) at each layer (analogous to each optimization step), based on multi-view image tokens (serving as gradient-like signals). This design allows the model to iteratively update the scene representation at every layer based on feedback from the multi-view input images, effectively mimicking the optimization process within a feed-forward architecture. Through this process, the learned neural network jointly examines the input view images and the evolving scene representation to identify where and how to make targeted updates that improve reconstruction quality.

Another core design principle of our approach is to decouple the representation, later transformed into 3D Gaussians, from direct dependence on input images, addressing the computational complexity and redundancy that arise in architectures that generate pixel-aligned Gaussians directly from multi-view inputs. By decoupling the representation and input images, we can use low-resolution representations to produce a compact set of Gaussians while still leveraging high-resolution input images for detailed guidance.

In addition, we propose an efficient mechanism for modeling the interaction between the representations and the input images. A naïve approach would involve computing full attention between all tokens across views, which quickly becomes computationally prohibitive. To overcome this, we initialize our representation using viewpoint embeddings, each tied to a specific input view. Interaction modeling is then split into two stages. First, we perform cross-attention between each viewpoint embedding and its corresponding image, which is highly efficient due to the one-to-one mapping. Next, we apply self-attention across all viewpoint embeddings. Importantly, since this second stage operates over a low-resolution representation space, it remains computationally tractable while facilitating global information exchange across views. Overall, this scalable design significantly reduces computational and memory overhead and allows for the incorporation of more viewpoints, thereby improving reconstruction fidelity.

We comprehensively evaluate the proposed method on large-scale datasets, RealEstate10K~\cite{zhou2018re10k} and DL3DV~\cite{ling2024dl3dv}. The experimental results demonstrate that \textit{iLRM} achieves superior rendering quality while substantially reducing both computational and memory overhead compared to recently proposed feed-forward Gaussian models. Moreover, in high-resolution and wide-coverage settings (\textbf{540$\times$960, 32 views}), our method completes inference in \textbf{0.5 seconds}, achieving comparable performance to an optimization-based approach, which takes about \textbf{8 minutes}.

\section{Related Works}
\subsection{Feed-forward 3D Gaussian Splatting}
\vspace{-1mm}
Feed-forward 3D Gaussian Splatting~\citep{charatan2024pixelsplat, chen2025mvsplat, xu2025depthsplat, nam2025generative, zhang2025gs-lrm, szymanowicz2024splatter, tang2025lgm, xu2024grm} capitalizes on robust priors learned from extensive datasets to estimate Gaussian primitive parameters and synthesize novel view images using sparse input data.
PixelSplat~\cite{charatan2024pixelsplat} and LatentSplat~\cite{wewer2024latentsplat} predict Gaussians from image features using an epipolar line sampling method to enhance geometric accuracy, while MVSplat~\cite{chen2025mvsplat} and MVSGaussian~\cite{liu2024mvsgaussian} construct cost volumes through a plane-sweep stereo approach. In a further development, Flash3D~\cite{szymanowicz2024flash3d} and DepthSplat~\cite{xu2025depthsplat} introduce a pre-trained depth estimation model~\cite{piccinelli2024unidepth, depthanything}, which improves the robustness of the spatial positions of 3D Gaussians. 
In contrast, GS-LRM~\cite{zhang2025gs-lrm} and Long-LRM~\cite{ziwen2025llrm} minimize reliance on explicit 3D priors by leveraging large-scale data-driven priors.

While demonstrating strong results, a major limitation of all the aforementioned approaches lies in their non-scalable architectural design, which restricts their ability to effectively leverage a large number of input views. Moreover, the one-shot generation strategy, which produces 3D representations in a single forward pass, often fails to capture complex geometric details and fine 3D consistency, making them suboptimal for high-quality 3D reconstruction. We address these limitations by proposing an iterative 3D reconstruction framework and scalable architectural designs.

\textbf{Iterative refinements.} Our work is also closely related to recent methods that adopt iterative refinement strategies, such as G3R~\cite{chen2024g3r} and Gen-Den~\cite{nam2025generative}. Both utilize actual gradients to update their representations more precisely. While promising, these approaches require additional computational burden for rendering multiple images per training iteration, and relying solely on gradients may risk overlooking valuable information present in the raw input images. Nonetheless, exploring how to incorporate gradient information remains an interesting direction for future work.

\subsection{3D Representations from Embeddings}
\vspace{-1mm}
Inspired by previous generative approaches~\cite{goodfellow2014generative, karras2019style, chan2022efficient}, recent works~\cite{hong2023lrm, chen2024lara, flynn2024quark} have investigated the synthesis of 3D representations directly from learnable embeddings, guided by input image supervision. This paradigm leverages the expressive capacity of latent spaces to encode rich geometric priors, which act as structural templates that guide the reconstruction process. Such approaches offer notable flexibility, allowing rendering from arbitrary viewpoints and adaptation to varying space scales and camera poses. However, both LRM~\cite{hong2023lrm} and Lara~\cite{chen2024lara} are limited to object-centric representations, restricting scalability to complex scenes involving multiple objects or large spatial layouts. The recently proposed Quark~\cite{flynn2024quark} also utilizes learnable embeddings to fuse visual cues from multiple images, demonstrating compelling results, but its representation is confined to the target view~\cite{xu2024murf, liu2024mvsgaussian, jin2024lvsm}, lacking an explicit and persistent 3D reconstruction.

In contrast to previous works, we construct scene-level explicit 3D representations from viewpoint embeddings by decoupling the generation of Gaussians from the input images. This separation enables iterative refinement of the embeddings using low-level visual features and provides flexible control over the density of the 3D representation, independent of the input image resolution.

\begin{figure*}[t]
    \centering
    \includegraphics[width=1.0\linewidth]{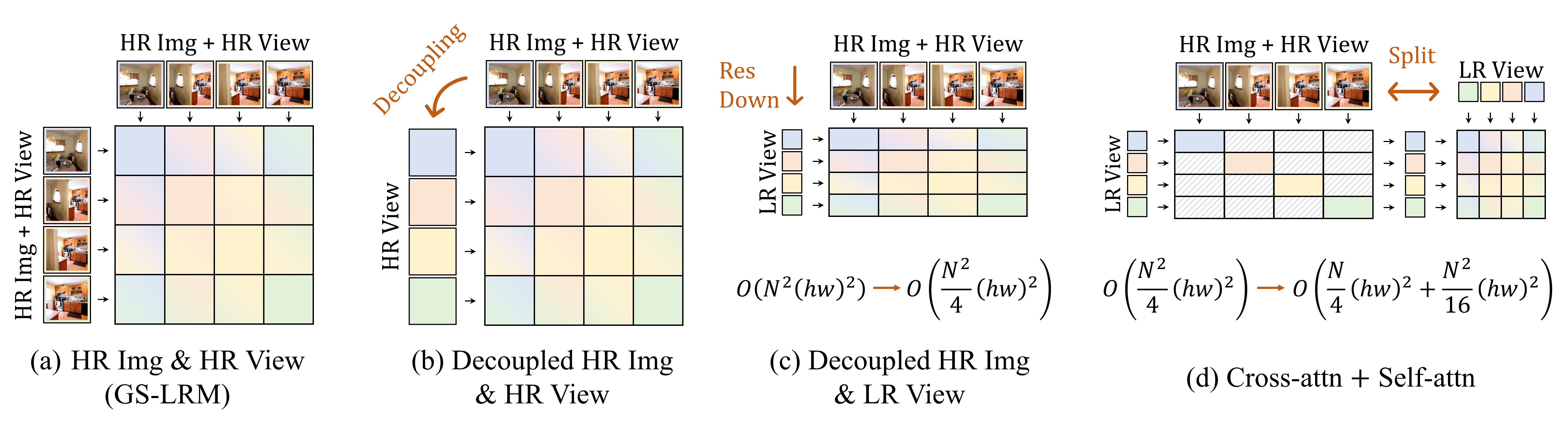}
    \vspace{-6mm}
    \caption{The proposed scalable architectural designs by decoupling viewpoint and image tokens, and modeling the global interactions via cross- and self-attentions ($N$: \# views, $h =H/p, w = W/p$).}
    \label{fig:eff_attn}
    \vspace{-3mm}
\end{figure*}

\section{Method}
\subsection{Motivation and Problem Statement}
\vspace{-1mm}
\label{subsec:motivation}

Existing generalizable 3D Gaussian reconstruction methods process multi-view images in an end-to-end fashion, often employing techniques such as epipolar line sampling~\cite{charatan2024pixelsplat}, plane-sweep stereo~\cite{chen2025mvsplat, chen2024mvsplat360, xu2025depthsplat}, or full-resolution attention~\cite{zhang2025gs-lrm, xu2024grm, ziwen2025llrm} to enforce multi-view consistency. While effective, these strategies introduce significant computational and memory overhead, limiting their scalability.

To address these challenges, we propose \textit{iLRM}, a novel feed-forward 3D reconstruction framework that decouples Gaussian generation from direct dependence on input images. Instead of generating pixel-aligned Gaussians, \textit{iLRM} initializes viewpoint-centric embeddings as the basis for constructing the 3D scene. These embeddings are then iteratively refined via cross-attention with multi-view image features, enabling the model to efficiently fuse geometric and appearance cues across views.

We start with $N$ multi-view images $\{I_i\}_{i=1}^N$ and 
camera poses $\{C_i\}_{i=1}^N$.
Based on this setup, our goal is to train a model $f_\theta$ that maps a set of viewpoints to 3D Gaussians, leveraging the associated multi-view images as visual cues throughout the reconstruction pipeline. More formally,
\vspace{-2mm}
\begin{equation}
    f_{\theta}: \{({C}_{i}, I_i) \}_{i=1}^{N} \mapsto \{({\mu}_k, \alpha_k, {\Sigma}_k, {c}_k )\}^{H^v W^v N}_{k=1},
\vspace{-2mm}    
\end{equation}
where $f_\theta$ is modeled as a feed-forward network with the model parameter $\theta$. $\mu_k, \alpha_k, \Sigma_k, c_k$ are attributes of 3D Gaussians, representing the mean, opacity, covariance, and color, respectively, while $H^v$ and $W^v$ denote the height and width of the generated Gaussians for each camera viewpoint. It is important to note that they do not correspond to the resolution of the input images.
We train our model using held-out target images along with their corresponding camera poses, enabling high-quality novel view synthesis.

\subsection{Architectural Design}
\vspace{-1mm}
\label{subsec:architecture}
We propose an end-to-end transformer that directly regresses 3D Gaussian parameters from viewpoint embeddings. To compensate for the absence of direct image input, we enrich these embeddings at each layer via cross-attention with multi-view image features. The resulting embeddings are further refined through self-attention to capture global dependencies across viewpoints.

\noindent\textbf{Viewpoint tokenization.}
Following previous works~\cite{tang2025lgm, zhang2025gs-lrm, jin2024lvsm}, we employ a Plücker ray embedding for each input view using the camera poses. Specifically, given the intrinsic, rotation, and translation, we construct the Plücker ray embeddings for each viewpoint. We then divide these viewpoint embeddings into non-overlapping patches of size $p \times p$, and reshape each patch into a 1D vector, resulting in a tensor of shape $H^v W^v / p^2 \times 6p^2$. Then, we encode it using a single linear layer to produce viewpoint tokens, $V_i^{(0)} \in \mathbb{R}^{H^v W^v / p^2 \times d}$. Plücker coordinates inherently capture spatial variations across pixels and views, allowing them to effectively differentiate between patches. As a result, we do not utilize additional positional embeddings.

\noindent\textbf{Multi-view image tokenization.}
For each input view image, which provides visual guidance to the reconstruction process, we extract both image features and corresponding pose information. Specifically, we divide an input image into non-overlapping patches and obtain two sets: RGB image patches and Plücker ray patches. These are then concatenated and linearly projected to construct the image patch tokens, $S_i \in \mathbb{R}^{HW / p^2 \times d}$,
\vspace{-1.2mm}
\begin{equation}
    S_{ij} = \text{Linear}(\text{concat}(I_{ij}, P_{ij})) \in \mathbb{R}^d,
    \vspace{-1.2mm}
\end{equation}
where $I_{ij} \in \mathbb{R}^{3p^2}, P_{ij} \in \mathbb{R}^{6p^2}$ represent the flattened $j$-th image and ray patches for the $i$-th view, respectively, and $HW / p^2$ is the number of tokens for each input view image.

\noindent\textbf{Scalable multi-view context modeling.}
Fig.~\ref{fig:eff_attn}-(a) shows the typical feed-forward 3D methods~\cite{zhang2025gs-lrm, chen2025mvsplat, xu2025depthsplat} using transformer architecture, which perform full attention across multi-view images, incurring a quadratic increase in computational cost with respect to both the number of views and the image resolution. Fig.~\ref{fig:eff_attn}-(b) depicts our decoupling approach.
Thanks to the decoupling technique, we can reduce the viewpoint resolution while still leveraging high-resolution multi-view images (Fig.~\ref{fig:eff_attn}-(c)).
We further decrease the computation cost by two-stage multi-view context modeling, per-view cross-attention and viewpoint self-attention (Fig.~\ref{fig:eff_attn}-(d)).
For example, given 16 input images with a resolution of 256~$\times$~256 and a patch size of 8, the relative computational cost follows the ratio (1:1:0.25:0.08, Fig.~\ref{fig:eff_attn}-(a):(b):(c):(d)), highlighting that our proposed method can accommodate more input views with significantly less computational burden.

\begin{figure}[!h]
    \centering
    \includegraphics[width=0.85\linewidth]{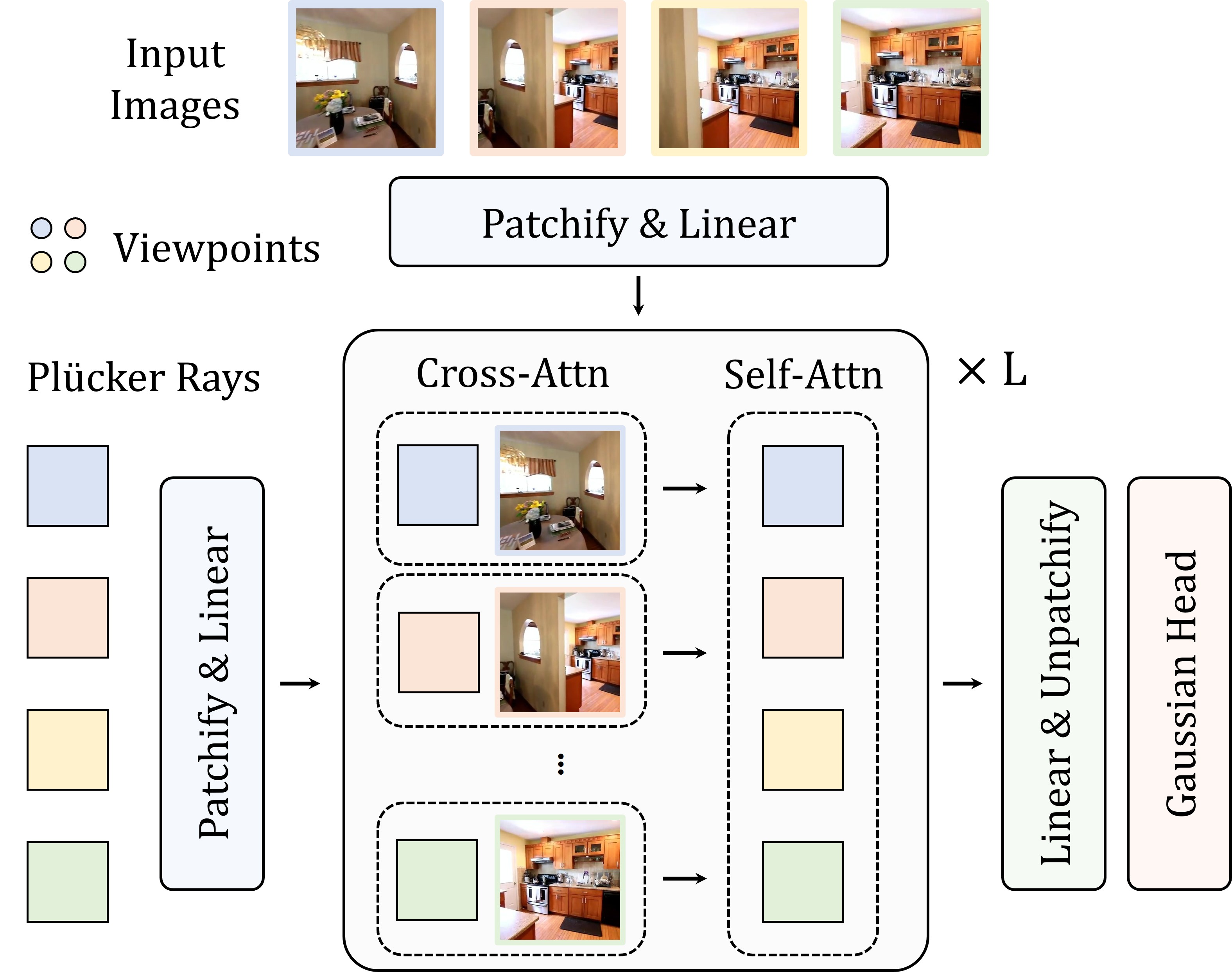}
    \vspace{-2mm}
    \caption{Network architecture.}
    \label{fig:update_layer}
    \vspace{-2mm}
\end{figure}

\begin{figure*}[t]
    \centering
    \includegraphics[width=1.0\linewidth]{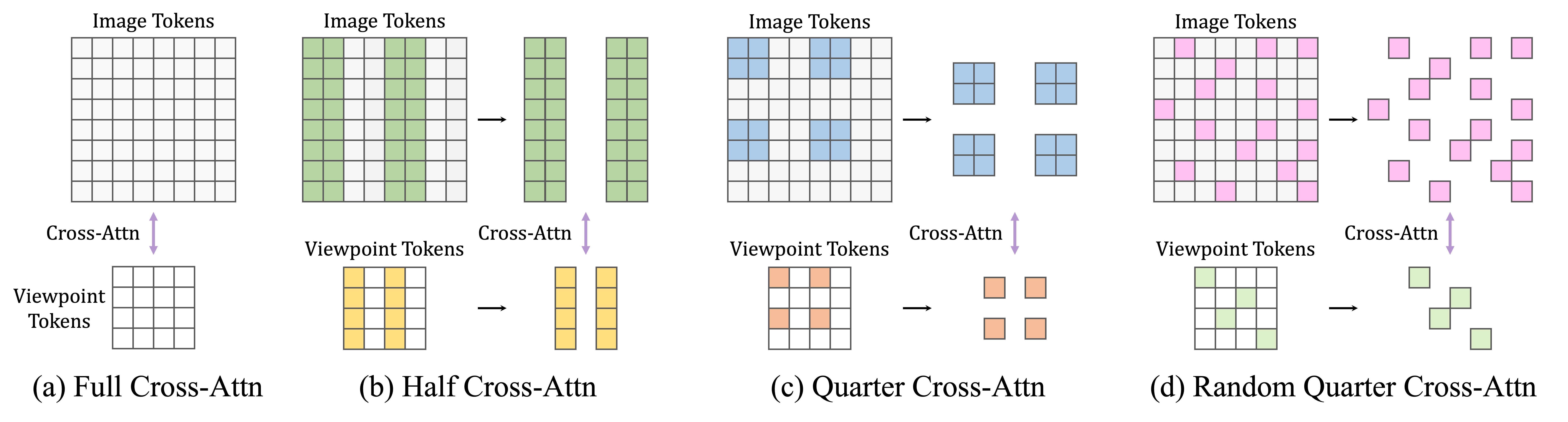}
    \vspace{-7mm}
    \caption{Various mini-batch cross-attention schemes. We primarily adopt ``Quarter Cross-attention" in our experiments.}
    \label{fig:cross_attn}
    \vspace{-3mm}
\end{figure*}

\noindent\textbf{Update block.}
Given a set of viewpoint tokens, we formulate the problem as an iterative refinement process, where the viewpoint tokens are progressively updated through interactions with multi-view image tokens, ultimately leading to more accurate and spatially consistent 3D Gaussian Splatting.
As shown in Fig.~\ref{fig:update_layer}, our model consists of multiple transformer modules, each comprising one cross-attention layer followed by one self-attention layer.
\vspace{-1.2mm}
\begin{align}
    \tilde{V}_i^{(l-1)} = \text{cross-attn}^{(l)}(V_i^{(l-1)}, S_i), \\
    \{V_i^{(l)} \}_{i=1}^N = \text{self-attn}^{(l)}( \{ \tilde{V}_i^{(l-1)} \}_{i=1}^N ),
    \vspace{-1.2mm}    
\end{align}
where the superscript $(l)$ denotes the layer index. In the cross-attention layers, the viewpoint tokens are refined by the visual information from their corresponding image tokens.
In the self-attention layers, the viewpoint tokens interact with each other to enhance their representations using global contextual information. Note that we use separate model parameters for the update blocks at different layers.

\noindent\textbf{Token uplifting.}
Standard cross-attention is typically applied between token sets of similar spatial resolutions. In our setting, however, we intentionally use lower-resolution (LR) viewpoint tokens compared to HR image tokens to improve scalability and efficiency, which may limit their ability to fully capture rich visual information. To bridge this gap, we propose a token uplifting strategy. Each LR viewpoint token is lifted by a linear query layer that expands its feature dimension by a factor of $k$, yielding a tensor of shape $H^v W^v / p^2 \times dk$, which is then reshaped to $H^v W^v k / p^2 \times d$ so that each original token corresponds to $k$ finer-grained query tokens for better visual correspondence during cross-attention.
After cross-attention with HR image tokens as keys and values, the resulting tensor is reshaped back to $H^v W^v / p^2 \times dk$ and projected to the original dimension $d$ via a linear layer, yielding refined viewpoint tokens of shape $H^v W^v / p^2 \times d$. 
We set $k = 2$ to balance representational capacity and efficiency.

\noindent\textbf{Mini-batch cross-attention.} 
In our architecture, viewpoint tokens are iteratively updated at each layer based on the image tokens via cross-attention. The proposed decoupling design allows us to arbitrarily reduce the number of viewpoint tokens for improved scalability, whereas the resolution of image tokens remains fixed due to their spatial nature. Consequently, the primary computational bottleneck in cross-attention lies in the high-resolution image tokens.

To address this, we propose several efficient cross-attention schemes, as illustrated in Fig.~\ref{fig:cross_attn}, aimed at improving scalability without sacrificing performance.
Our design is conceptually inspired by mini-batch gradient descent in optimization, where only a subset of data points is sampled in each iteration to reduce computational cost.
Similarly, our mechanism selectively samples subsets of both image tokens and viewpoint tokens during cross-attention.
While random token sampling (Fig.~\ref{fig:cross_attn}-(d)) is ideal in theory, it complicates efficient implementation. To mitigate this, we design structured sampling strategies that are simple to implement and demonstrate strong empirical performance.

\noindent\textbf{Decoding into 3D Gaussians.}
After the final self-attention layer, we decode the final layer's viewpoint tokens, $V_i^{(L)}$, into Gaussian parameters through a single linear layer and apply post-activation functions. For a detailed description, please refer to our supplementary materials.

\noindent\textbf{Interpretation.}
Compared to standard self-attention, $S^{(l)} = S^{(l-1)} + f(S^{(l-1)})$,
our method applies evidence-conditioned updates, $V^{(l)} = V^{(l-1)} + F(V^{(l-1)}, S)$,
where the image tokens $S$ are fixed and provide detailed visual guidance.
This resembles a gradient descent iteration,
$V^{(l)} \approx V^{(l-1)} - \eta \nabla_V \mathcal{E}(V^{(l-1)}; S)$,
where $\mathcal{E}$ is an implicit objective function, making each layer a feedback correction step rather than a pure feature transformation.
Our mini-batch variant further extends this view as
$V_{\mathrm{mb}}^{(l)} \approx V_{\mathrm{mb}}^{(l-1)} - \eta \nabla_{V_{\mathrm{mb}}} \mathcal{E}(V_{\mathrm{mb}}^{(l-1)}; S_{\mathrm{mb}})$,
where $V_{\mathrm{mb}}$ and $S_{\mathrm{mb}}$ denote a subset of viewpoint tokens and their corresponding image tokens, respectively.

\subsection{Training Objectives}
\vspace{-1mm}

\label{subsec:objective}
We rasterize 3D Gaussians from viewpoint tokens to obtain rendered images $\hat{I}_t$, supervised against ground-truth images $I_t$ via MSE and perceptual loss~\cite{chen2017photographic, li2020crowdsampling}:
\begin{align}
\mathcal{L}_\text{total} = \sum_{t \in \mathcal{T}}\mathcal{L}_\text{MSE}(\hat{I}_t, I_t)+\lambda\mathcal{L}_\text{perceptual}(\hat{I}_t, I_t),
\end{align}
where $\mathcal{T}$ is the set of target view indices and $\lambda{=}0.5$ balances the two loss terms.

\section{Experiments}
\vspace{-1mm}
\noindent\textbf{Datasets.} We train our model on two large-scale datasets: RealEstate10K (RE10K)~\cite{zhou2018re10k} and DL3DV~\cite{ling2024dl3dv}, and evaluate it on three datasets, including ACID~\cite{liu2021acid}. We adopt the RE10K split following~\cite{charatan2024pixelsplat} and the official split for DL3DV.
We use an image resolution of 256$\times$256 for the RE10K and ACID datasets, while for the DL3DV dataset, we use a resolution of 256$\times$448 and 512$\times$960. In addition, we employ the undistorted version of the DL3DV dataset at a resolution of 540$\times$960, which originates from Long-LRM~\cite{ziwen2025llrm}.

\noindent\textbf{Implementation details.}
Our model consists of 12 update layers, each containing one cross-attention and one self-attention block. Inside each attention module, we adopt a pre-normalization method with LayerNorm~\cite{ba2016layernorm} and QK-Norm~\cite{henry2020qknorm} method with an RMSNorm~\citep{zhang2019rmsnorm} layer. Also, each block utilizes 12-head attention~\cite{vaswani2017attention} and two GELU~\cite{hendrycks2016gelu}-activated linear layers. We set the hidden dimension to $d$ = 768, and use a patch size of $p$ = 8.  For more details, please refer to the supplementary material.

\noindent\textbf{Evaluation.} We compare our model against recent generalizable 3D reconstruction methods~\cite{charatan2024pixelsplat, chen2025mvsplat, zhang2025gs-lrm, xu2025depthsplat, nam2025generative, ziwen2025llrm} as well as optimization-based approach~\cite{kerbl20233dgs}. For evaluation, we follow the settings from~\cite{charatan2024pixelsplat, ye2024no} for RE10K and \cite{xu2025depthsplat, ziwen2025llrm} for DL3DV. We denote our various viewpoint settings as $(V, H/F, F)$, where $V$ is the number of viewpoints, and the following entries indicate the resolutions of viewpoint and image tokens (\textit{H}: half-resolution, \textit{F}: full-resolution). For example, a setting of $(2, H, F)$ indicates two viewpoints tokens with half-resolution and full-resolution image tokens. $MC$ refers to our quarter mini-batch cross-attention (Fig.~\ref{fig:cross_attn}-(c)). Note that our 2-view full-resolution setting $(2, F, F)$ does not include token uplifting, as resolutions are identical. Additionally, when using more views than is required in the evaluation protocol, we sample extra views, ensuring there is no overlap with the target indices.
\vspace{-1mm}

\begin{table}[!h]
    \centering
    \resizebox{1.0\columnwidth}{!}{
        \setlength{\tabcolsep}{2pt}    
        \begin{tabular}{@{}lcccccc@{}}
        \toprule
        Method & \#Param (M) & PSNR $\uparrow$ & SSIM $\uparrow$ & LPIPS $\downarrow$ & \# Gaussians & Time (s)\\
        \midrule
        pixelSplat & 125 & 25.89 & 0.858 & 0.142 & 131,072 & 0.101 \\
        MVSplat & 12 & 26.39 & 0.869 & 0.128 & 131,072 & 0.047 \\
        GS-LRM* & 300 & 28.10 & 0.892 & 0.114 & 131,072 & - \\
        DepthSplat & 354 & 27.47 & 0.889 & 0.114 & 131,072 & 0.065 \\ 
        Gen-Den & 28 & 27.08 & 0.879 & 0.120 & 347,072 & 0.224 \\
        Ours $(2, F, F)$ & 171 & \best{28.65} & \best{0.900} & \best{0.110} & 131,072 & \best{0.025} \\
        \midrule
        Ours $(4, H, F)$ & 185 & 30.37 & 0.923 & 0.095 & 65,536 & 0.027 \\ 
        Ours-MC $(4, H, F)$ & 185 & 30.10 & 0.919 & 0.098 & 65,536 & 0.027 \\
        Ours $(8, H, F)$ & 185 & 31.57 & 0.935 & 0.082 & 131,072 & 0.028 \\
        Ours-MC $(8, H, F)$ & 185 & 31.24 & 0.933 & 0.084 & 131,072 & 0.029 \\    
        \bottomrule
        \end{tabular}
    }
    \vspace{-3mm}    
    \caption{
    Quantitative comparisons on the RE10K dataset with various view configurations. Inference time is measured on a RTX 4090 GPU. * indicates closed-source methods. The time difference in the MC variant is negligible due to the short sequence length in inference. For a more comprehensive analysis of our mini-batch cross-attention, see Tab.~\ref{tab:mini-batch on re10k}.
    }        
    \label{tab:quantitative result on re10k}
    \vspace{-4mm}
\end{table}

\begin{figure}[!h]
    \centering
    \includegraphics[width=1.0\linewidth]{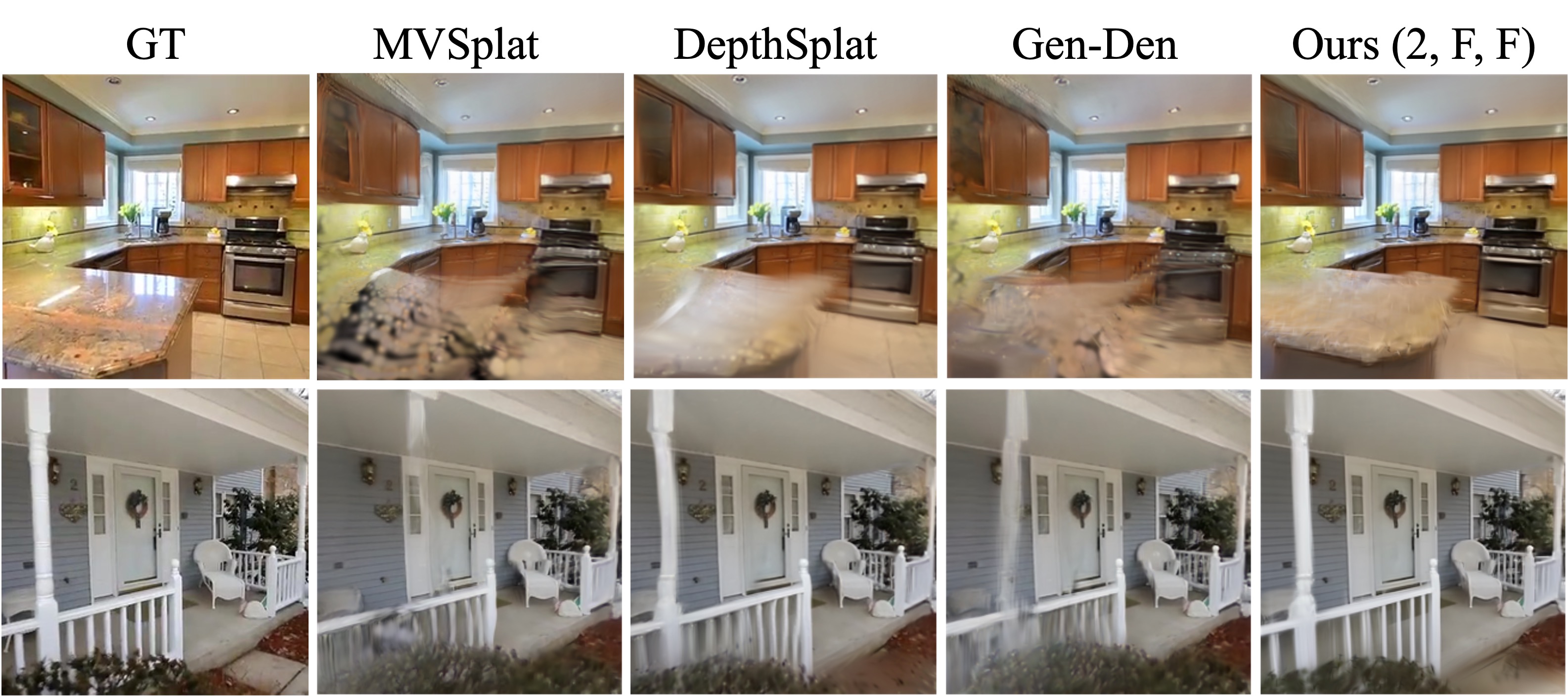}
    \vspace{-7mm}        
    \caption{Qualitative comparison on the RE10K dataset.}
    \label{fig:qual_re10k}
    \vspace{-2mm}    
\end{figure}

\begin{table}[!h]
    \centering
    \resizebox{1.0\columnwidth}{!}{
        \begin{tabular}{@{}lcccccc@{}}
        \toprule
        \multirow{2}{*}{Method} & \multicolumn{3}{c}{ACID} & \multicolumn{3}{c}{DL3DV} \\
        \cmidrule(lr){2-4}\cmidrule(lr){5-7}
         & PSNR $\uparrow$ & SSIM $\uparrow$ & LPIPS $\downarrow$ & PSNR $\uparrow$ & SSIM $\uparrow$ & LPIPS $\downarrow$ \\
        \midrule
        MVSplat & 28.15 & 0.841 & 0.147 & 22.65 & 0.737 & 0.191 \\
        DepthSplat & 28.37 & 0.847 & \best{0.141} & 24.28 & 0.813 & 0.147 \\
        Gen-Den & 28.61 & 0.847 & \best{0.141} & 22.92 & 0.750 & 0.188 \\
        Ours $(2, F, F)$ & \best{29.24} & \best{0.856} & 0.143 & \best{25.35} & \best{0.826} & \best{0.144} \\
        \midrule
        Ours $(4, H, F)$ & 30.10 & 0.877 & 0.138 & 27.90 & 0.877 & 0.122 \\
        Ours-MC $(4, H, F)$ & 29.90 & 0.873 & 0.141 & 27.68 & 0.881 & 0.127 \\        
        Ours $(8, H, F)$ & 30.96 & 0.894 & 0.122 & 29.56 & 0.907 & 0.101 \\        
        Ours-MC $(8, H, F)$ & 30.72 & 0.890 & 0.125 & 29.33 & 0.904 & 0.102 \\        
        \bottomrule
        \end{tabular}
    }
    \vspace{-3mm}    
    \caption{
    Cross-dataset generalization on the ACID and DL3DV (256$\times$256) using a model trained on the RE10K dataset.
    }            
    \label{tab:cross-dataset generalization on scene}
    \vspace{-2mm}
\end{table}

\begin{table}[!h]
    \centering
    \resizebox{1.0\columnwidth}{!}{
        \setlength{\tabcolsep}{2pt}       
        \begin{tabular}{@{}lccccccc@{}}
        \toprule
        Method & Views & PSNR $\uparrow$ & SSIM $\uparrow$ & LPIPS $\downarrow$ & \# Gaussians & Time (s) & Memory (GB) \\
        \midrule

        MVSplat & {6} & 22.93 & 0.775 & 0.193 & 688,128  & 0.279 & 5.87 \\

        \midrule

        \multirow{3}{*}{DepthSplat} & 6 & 24.19 & 0.823 & \best{0.147} & 688,128 & 0.102 & 3.55 \\        
        & 11 & 24.28 & 0.833 & 0.141 & 1,261,568 & 0.170 & 6.01 \\
        & 24 & 22.37 & 0.781 & 0.195 & 2,752,512 & 0.371& 12.39 \\ 
        
        \midrule
        
        \multirow{3}{*}{Ours} & $(6, H, F)$ & \best{25.60} & \best{0.830} & 0.168 & \best{172,032} & \best{0.031} & \best{1.40} \\        
        & $(11, H, F)$ & \best{26.99} & \best{0.865} & \best{0.140} & \best{315,392} & \best{0.051} & \best{1.59} \\
        & $(24, H, F)$ & \best{27.38} & \best{0.882} & \best{0.126} & \best{688,128} & \best{0.123} & \best{2.01} \\        
        \bottomrule
        \end{tabular}
    }
    \vspace{-3mm}    
    \caption{
    Quantitative comparisons on the DL3DV dataset under the 50-frame baseline setting (256$\times$448). Inference time and memory consumption are measured on a RTX 4090 GPU.}
    \label{tab:quantitative result on dl3dv}  
    \vspace{-3mm}
\end{table}

\begin{figure}[!h]
    \centering
    \includegraphics[width=1.0\linewidth]{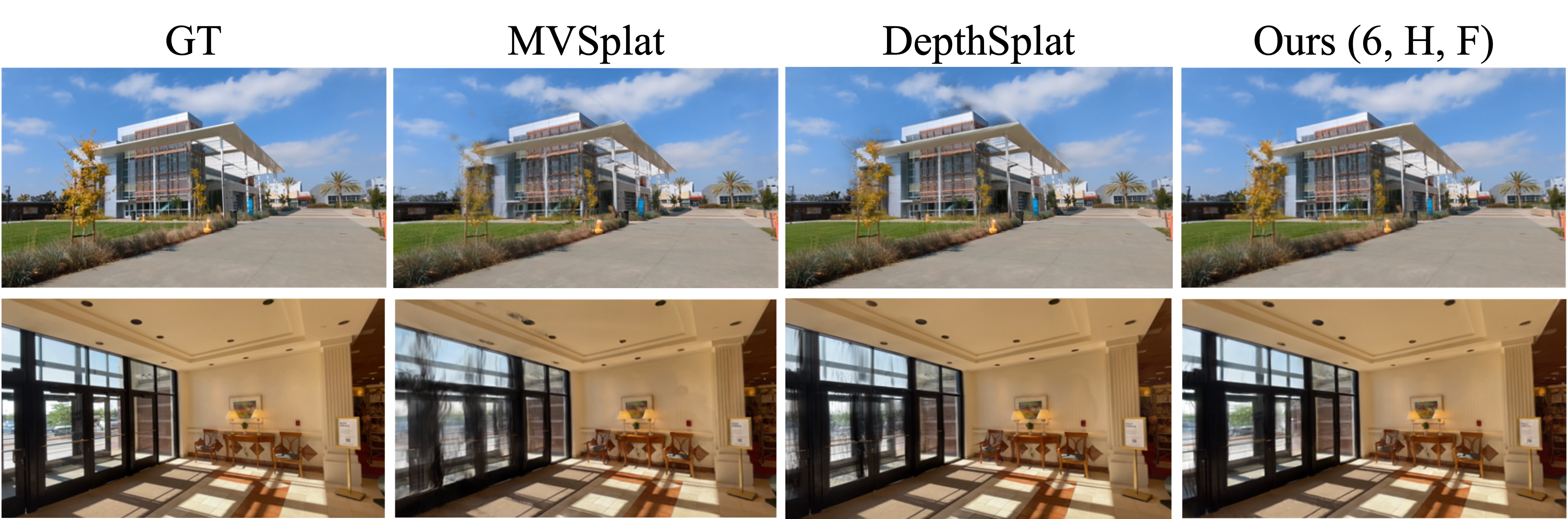}    
    \vspace{-7mm}    
    \caption{Qualitative comparison on DL3DV dataset (256$\times$448).}    
    \label{fig:qual_dl3dv}
    \vspace{-6mm}    
\end{figure}


\begin{figure*}[!t]
    \centering
    \includegraphics[width=1.0\linewidth]{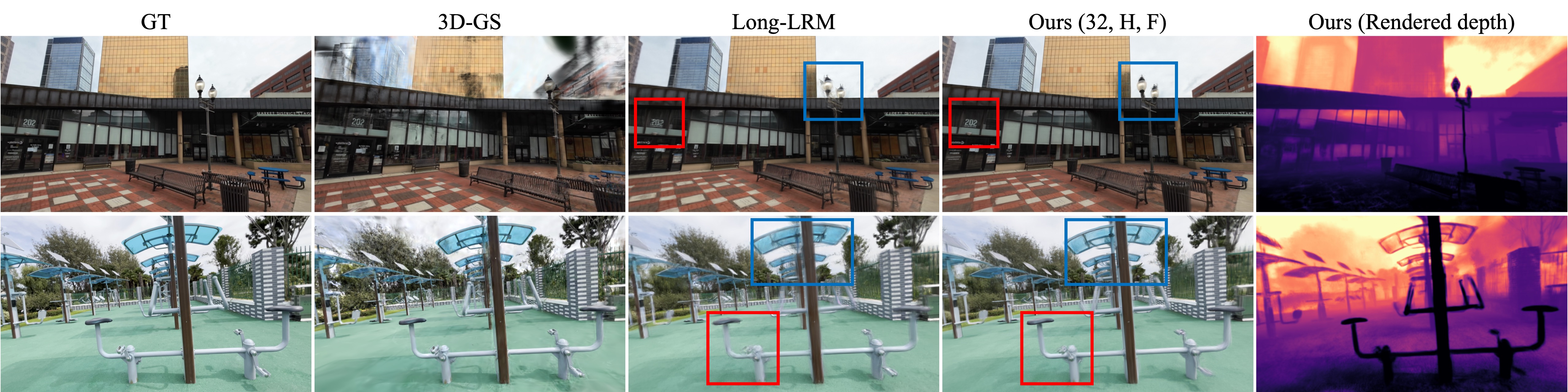}
    \vspace{-7mm}
    \caption{Qualitative comparison on undistorted DL3DV dataset under the wide-baseline setting (32 input images, 540$\times$960, zero-shot).}    
    \label{fig:qual_hr_dl3dv2}
    \vspace{-3mm}    
\end{figure*}

\begin{figure*}[!t]
    \centering
    \includegraphics[width=1.0\linewidth]{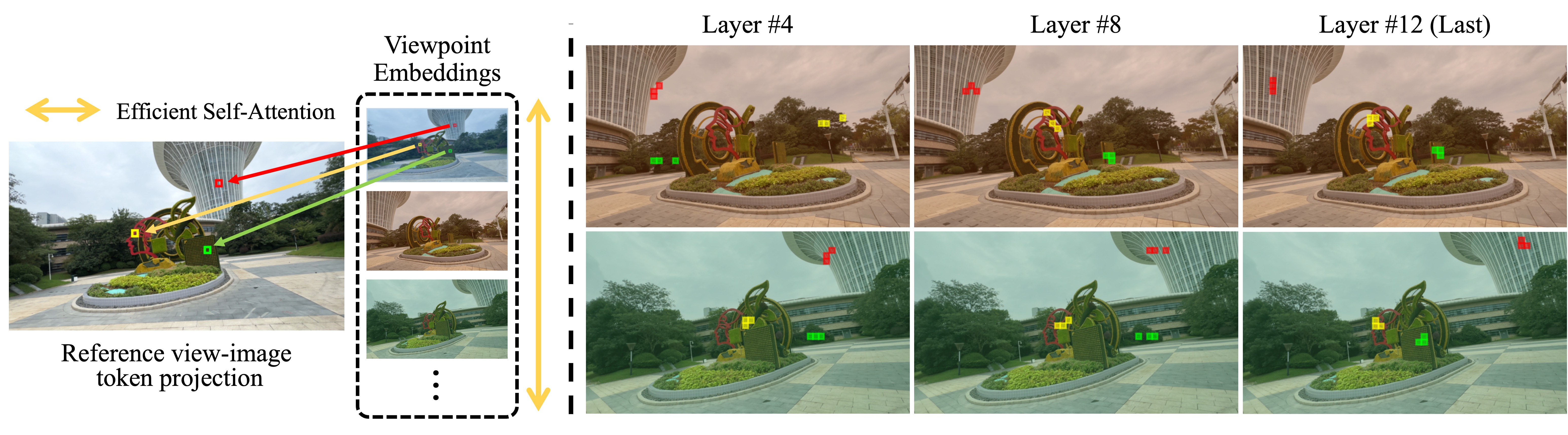}
    \vspace{-8mm}
    \caption{For the colored query patches in the reference viewpoint (\textcolor[HTML]{C02A1E}{red}, \textcolor[HTML]{D6D659}{yellow}, \textcolor[HTML]{62CF42}{green}), we visualize top-3 attended tokens from other viewpoints throughout the iterative refinement process.}    
    \label{fig:attention}
    \vspace{-4mm}    
\end{figure*}

\subsection{Results}
\vspace{-1mm}
In Tab.~\ref{tab:quantitative result on re10k},~\ref{tab:cross-dataset generalization on scene} and Fig.~\ref{fig:qual_re10k}, we compare our approach with feed-forward methods on the RE10K dataset and cross-dataset generalization on ACID and DL3DV. Furthermore, we report results with an increased number of input views (4 and 8), which incur less than half of the computation time compared to the baseline (DepthSplat) while achieving superior performance. For the DL3DV dataset, our method consistently outperforms the baselines across various viewpoint and resolution configurations, including inference speed and memory usage, while achieving efficient scene representation with fewer Gaussians, as shown in Tab.~\ref{tab:quantitative result on dl3dv},~\ref{tab:hr_table} and Fig.~\ref{fig:qual_dl3dv}. While DepthSplat and our method are both trained under varying numbers of input views, our approach demonstrates enhanced scalability with respect to the increasing number of views in Tab.~\ref{tab:quantitative result on dl3dv}.

\begin{table}[!h]
    \centering
    \resizebox{1.0\columnwidth}{!}{
        \setlength{\tabcolsep}{2pt}       
        \begin{tabular}{@{}lccccccc@{}}
        \toprule
        Method & Views & PSNR $\uparrow$ & SSIM $\uparrow$ & LPIPS $\downarrow$  & \# Gaussians & Time (s) & Memory (GB) \\
        \midrule       
        DepthSplat & 12 & 21.38 & 0.739 & 0.265 & 5,898,240 & -  & OOM \\       
        Ours & $(12, H, F)$ & \best{24.35} & \best{0.781} & \best{0.256} & \best{1,474,560} & \best{0.415}  & \best{3.53} \\       
        \bottomrule
        \end{tabular}
    }
    \vspace{-2mm}        
    \caption{
    Quantitative comparisons on the DL3DV dataset under the 100-frame baseline setting (512$\times$960). Inference time and memory consumption are measured on a single RTX 4090 GPU. Since DepthSplat encounters out-of-memory (OOM) issue on the device, we evaluate its performance using a H100 GPU.
    }
    \label{tab:hr_table}
    \vspace{-6mm}        
\end{table}

\begin{table}[!h]
    \centering
    \resizebox{1.0\columnwidth}{!}{
        \begin{tabular}{lccccc}
        \toprule
        Method & Views & Time $\downarrow$ & PSNR $\uparrow$ & SSIM $\uparrow$ & LPIPS $\downarrow$  \\
        \midrule
        3D-GS & 16 & 8min & 21.48 & 0.753 & 0.252  \\     
        \midrule
        Long-LRM & 16 & 0.50sec & 22.66 & 0.740 & \best{0.292}  \\
        Ours & $(16, H, F)$ & \best{0.19sec} & \best{22.91} & \best{0.766} & 0.295  \\        
        \midrule
        \midrule
        3D-GS & 32 & 8min & 24.43 & 0.827 & 0.191  \\     
        \midrule
        Long-LRM & 32 & 0.84sec & 23.97 & 0.778 & 0.267  \\
        Ours & $(32, H, F)$ & \best{0.53sec} & \best{24.30} & \best{0.803} & \best{0.256}  \\
        \midrule
        Long-LRM$_{10}$ & 32 & 11sec & 25.56 & 0.826 & 0.237  \\
        Ours$_{10}$ & $(32, H, F)$ & \best{4.5sec} & \best{25.67} & \best{0.844} & \best{0.230}  \\
        \midrule
        \midrule
        Long-LRM (Unseen) & 40 & 1.05sec & 24.18 & 0.787 & 0.260  \\
        Ours (Unseen) & $(40, H, F)$ & \best{0.76sec} & \best{24.54} & \best{0.811} & \best{0.248}  \\   
        \midrule
        Long-LRM (Unseen) & 48 & 1.38sec & 24.30 & 0.797 & 0.252  \\
        Ours (Unseen) & $(48, H, F)$ & \best{1.04sec} & \best{24.78} & \best{0.820} & \best{0.240}  \\  
        \bottomrule
        \end{tabular}
    }
    \vspace{-2mm}     
    \caption{
    Quantitative comparisons on the undistorted DL3DV dataset (540$\times$960). We utilized flash attention v3~\citep{shah2024flashattention} using a H100 GPU. We re-evaluate Long-LRM~\cite{ziwen2025llrm} with their official checkpoint except for 16-view metrics (16-view weights are not released).
    }      
    \label{tab:undisthr_table}   
    \vspace{-7mm}
\end{table}

In the wide-coverage setting (Tab.~\ref{tab:undisthr_table}), we evaluate performance using various numbers of high-resolution input images under full-frame coverage. For comparison, we also include optimization-based 3D-GS~\cite{kerbl20233dgs} trained for 30k iterations using the input images and camera poses. Long-LRM$_{10}$ means finetuning 10 epochs initialized from the Long-LRM’s generated Gaussians. Since our approach produces more compact 3D Gaussian representations (4$\times$ fewer), the finetuning process is significantly faster than the baseline. We further evaluate longer-context generalization ability (40 and 48 views) using a model trained with 32 views. Our method achieves better performance and faster inference across all metrics and scales more favorably with the number of views while maintaining compact scenes.

\begin{figure*}[h]
    \centering
    \resizebox{0.9\linewidth}{!}{    \begin{minipage}[h]{0.5\textwidth}
        \captionsetup{type=table}
        \setlength\tabcolsep{5pt}
        \centering
        \resizebox{0.9\textwidth}{!}{%
        \begin{tabular}{r|c|ccc} 
        \toprule
        & \# Params & PSNR \(\uparrow\) & SSIM \(\uparrow\)  & LPIPS \(\downarrow\) 
        \\ 
        \midrule
        12 layers (base) & 185M & \best{29.24} & \best{0.907} & \best{0.109} \\
        9 layers & 139M & 29.01 & 0.903 & 0.112 \\
        6 layers & 94M & 28.68 & 0.898 & 0.116 \\
        3 layers & 48M & 28.04 & 0.887 & 0.126 \\
        \bottomrule        
        \end{tabular}
        }
        \vspace{-2mm}         
        \caption{Ablations on model size.}        
        \label{table:ablation_model_size}
    \end{minipage}
    \hfill
    \begin{minipage}[h]{0.50\textwidth}
        \captionsetup{type=table}
        \setlength\tabcolsep{5pt}
        \centering
        \resizebox{0.9\textwidth}{!}{%
        \begin{tabular}{r|ccc} 
        \toprule
        & PSNR \(\uparrow\) & SSIM \(\uparrow\)  & LPIPS \(\downarrow\)  \\ 
        \midrule
        Baseline (12 layers) & \textbf{29.24} & \textbf{0.907} & \textbf{0.109} \\
        w/o iter. refinement  & 28.58 & 0.893 & 0.127  \\
        w/o resolution decoupling  & 28.47 & 0.891 & 0.123  \\
        w/o token uplifting & 28.90 & 0.901 & 0.113  \\ 
        \bottomrule
        \end{tabular}
        }
        \vspace{-2mm}         
        \caption{
        Ablations on model architecture.
        }                
        \label{table:ablation_model_arch}
    \end{minipage}}
    \vspace*{-5mm}
\end{figure*}

\subsection{Attention Visualization}
\vspace{-1mm}
We investigate how our method achieves global consistency throughout the iterative refinement process. Using the first input view as the reference, we select three query patches from its viewpoint embedding, and visualize top-3 attended tokens in the other viewpoint embeddings. For ease of visualization, we project the selected tokens onto the corresponding images via spatial upsampling. As shown in Fig.~\ref{fig:attention}, attended tokens from other viewpoints gradually shifts toward geometrically and semantically corresponding regions as the layers go deeper, demonstrating the progressive, iterative refinement of the multi-view scene representation, which aligns our proposed design motivation.

\subsection{Computational Costs of Training}
\vspace{-1mm}
We report detailed comparisons of computational costs during training in Tab.~\ref{tab:mini-batch on re10k}. The iteration time is measured under the same setting: half-resolution 8 viewpoints $(8, H, F)$, and a batch size of 16 on a single RTX 4090 GPU. For memory comparison, to provide a clearer analysis, all models are run without gradient checkpointing on a single H100 GPU. Lastly, we present a theoretical comparison of FLOPs that further underscores the efficiency of our method, with only a marginal drop in performance. For detailed calculations of FLOPs, please refer to our supplementary material.

\begin{table}[!h]
    \centering    
    \resizebox{1.0\columnwidth}{!}{
        \setlength{\tabcolsep}{2pt}     
        \begin{tabular}{@{}lcccccc@{}}
        \toprule
        Method & PSNR $\uparrow$ & SSIM $\uparrow$ & LPIPS $\downarrow$  & Iteration (s) & Memory (GB) & GFLOPs \\
        \midrule
        Baseline & \best{30.39} & \best{0.923} & \best{0.095} & 1.51 & 62.5  & 3.83 \\
        w/ Half Cross-attn & 30.25 & 0.922 & 0.096 & 1.13 & 47.4 & 1.71 \\        
        w/ Quarter Cross-attn & 30.08 & 0.919 & 0.098 & \best{0.94} & \best{39.0} & \best{0.81} \\        
        \bottomrule
        \end{tabular}
    }
    \vspace{-3mm}    
    \caption{
    Quantitative comparison of our mini-batch cross-attention on the RE10K dataset, with iteration time and memory consumption measured during training.
    }      
    \label{tab:mini-batch on re10k}     
    \vspace{-4mm}
\end{table}

\subsection{Ablations and Analysis}
\vspace{-1mm}
Tab.~\ref{table:ablation_model_size} presents the ablations on the number of update layers. All variants are trained under half-resolution 4 viewpoints setting $(4, H, F)$, on the RE10K dataset with batch size 16. The results demonstrate consistent performance gains as the number of layers increases. From the perspective of the iterative refinement procedure, increasing the number of layers can be interpreted as introducing more optimization steps, which aligns with our intuition that deeper refinement leads to more accurate 3D representations.

In Tab.~\ref{table:ablation_model_arch}, we report the ablation results on architectural components. All experiments follow the model-size ablation training setup with a 12-layer baseline. More extensive ablation studies are provided in the supplementary material.

\noindent\textbf{1) Iterative refinement.} The cross-attention blocks in our model keep providing visual evidence (image) into the viewpoint tokens as part of the iterative refinement process. We validate this by replacing per-layer cross-attention with a single cross-attention in the first layer: the baseline has 12 layers (each with cross- then self-attention), while the variant has 1 cross-attention followed by 23 self-attention layers. The result shows that our consecutive cross-attention with image features plays a critical role in refining the viewpoint embeddings especially in terms of the LPIPS metric.

\noindent\textbf{2) Resolution decoupling.} Our design decouples image resolution from the viewpoint representation, so cross-attention consumes high-resolution image features while the scene tokens remain lightweight. When image features are constrained to the viewpoint resolution (prior approaches~\cite{zhang2025gs-lrm, ziwen2025llrm}), performance drops, indicating that resolution decoupling is essential for simultaneously preserving compactness and high-fidelity reconstructions.

\noindent\textbf{3) Token uplifting.} Removing the token uplifting mechanism leads to a drop in performance across all metrics compared to baseline. This validates the importance of expanding low-resolution view tokens before cross-attention with high-resolution image tokens. Without this step, the model struggles to capture fine-grained spatial correspondences, resulting in a degraded reconstruction quality.

\vspace{-1.2mm}
\section{Limitations}
\vspace{-1.5mm}
One limitation of this work is the self-attention bottleneck across many input views.
While our compact viewpoint embeddings substantially reduce the computational cost,
challenges may arise as the number of input views increases considerably. In this study, aiming for scalable feed-forward 3D models, we present the first implementation of the framework that iteratively refines 3D representations by leveraging high-resolution image information at every layer. Further development of more scalable alternatives~\cite{child2019generating, mamba2} would be a valuable direction for future research.

A second limitation is the reliance on accurate camera poses~\cite{Schonberger_2016_CVPR_colmap, pan2024global} in static scenarios. Furthermore, because our primary goal is high fidelity novel view synthesis with rendering supervision, the recovered geometry may be less accurate than explicit geometry-supervised methods~\cite{wang2024dust3r, wang2025vggt}. Even so, our scalable and flexible structure provides a natural basis for relaxing these assumptions, as joint pose refinement~\cite{zhang2025flare} or even pose-free variants~\cite{ye2024no, hong2024pf3plat, kang2025selfsplat, jiang2025rayzer} could be realized by training on suitable datasets and supervisions without modifying the core architectural design.


\vspace{-1.2mm}
\section{Conclusion}
\vspace{-1.5mm}
In this work, we present an iterative Large 3D Reconstruction Model (\textit{iLRM}), a feed-forward architecture that reflects per-scene optimization-based schemes, by stacking multiple update layers composed of cross- and self-attention modules. By decoupling Gaussian representations from input images and splitting the update mechanism into per-view interactions with image features and global aggregation over compact viewpoint embeddings, \textit{iLRM} enables efficient, scalable, and high-quality 3D reconstruction across diverse scenes. We believe that \textit{iLRM} lays a strong foundation for future research in feed-forward 3D reconstruction.

\section*{Acknowledgements}
This work was supported by Samsung Research Funding \& Incubation Center of Samsung Electronics under Project Number SRFC-IT2401-01, the Artificial Intelligence Industrial Convergence Cluster Development Project funded by the Ministry of Science and ICT (MSIT, Korea) \& Gwangju Metropolitan City, and the Institute of Information \& Communications Technology Planning \& Evaluation (IITP) grant funded by the Korean government (MSIT) under the following projects: (No. RS-2024-00457882, AI Research Hub Project); (No. RS-2025-25441838, Development of a human foundation model for human-centric universal artificial intelligence and training of personnel); (No. RS-2020-II201361, Artificial Intelligence Graduate School Program (Yonsei University)); and (No. RS-2025-02653113, High-Performance Research AI Computing Infrastructure Support at the 2 PFLOPS Scale).

{
    \small
    \bibliographystyle{ieeenat_fullname}
    \bibliography{main}
}

\appendix

\twocolumn[
    \centering
    \Large
    \textbf{\thetitle}\\
    \vspace{0.5em}Supplementary Material \\
    \vspace{1.0em}
] %

\section{Additional Implementation Details}
\vspace{-2mm}
We initialize model weights using a zero-mean normal distribution with a standard deviation of 0.02. Bias terms are omitted in all Linear and normalization layers. The model is trained using the AdamW~\cite{loshchilov2017adamw} optimizer with hyperparameters $\beta_1 = 0.9$ and $\beta_2 = 0.95$. A weight decay of 0.05 is applied to all parameters except the weights of LayerNorm~\cite{ba2016layernorm}. We use a cosine learning rate schedule with a peak learning rate of 2e-4 and a warmup of 2500 iterations. Our training setup largely follows the configuration proposed in~\cite{zhang2025gs-lrm, jin2024lvsm}. 

For the RealEstate10K (RE10K)~\cite{zhou2018re10k} dataset, the 8-view half-resolution viewpoint setting \textit{(8, H, F)} is trained on 8 H100 GPUs with a total batch size of 256 for 50,000 iterations. Similarly, the 4-view half-resolution viewpoint setting \textit{(4, H, F)} is trained on 8 RTX 4090 GPUs with a total batch size of 128 for 100,000 iterations. The mini-batch cross-attention variants were also trained with the equivalent computational budgets for each viewpoint setting. Lastly, the 2-view full-resolution viewpoint setting \textit{(2, F, F)}, which serves as the reference point, is trained on 8 H100 GPUs for 200,000 iterations.

There are two variants in the DL3DV dataset~\cite{ling2024dl3dv}. \textbf{1)} For comparison with MVSplat~\cite{chen2025mvsplat} and DepthSplat~\cite{xu2025depthsplat}, we initialize from the pretrained \textit{(8, H, F)} model trained on the RE10K dataset, and finetune it on 8 H100 GPUs with a total batch size of 96 for 100,000 iterations for LR (256$\times$448), and additional 50,000 iterations for HR (512$\times$960). During training, the number of input viewpoints is randomly sampled between 6 and 11 to expose the model to varying numbers of viewpoints/images. Following this stage, the model is further finetuned under the high-resolution setting (512$\times$960). \textbf{2)} For comparison with LongLRM~\cite{ziwen2025llrm}, which incorporates an undistortion preprocessing step, we adopt the training protocol described in the original work, using 8 H200 GPUs. The training resolution is scheduled in a curriculum of 256$\times$256, 512$\times$512, and 540$\times$960. 

\noindent\textbf{Gaussian representations.} After the final self-attention layer, the viewpoint features are decoded into Gaussian parameters using a single linear layer with an output dimension of 16. The Gaussian positions, denoted as $\mu$, consist of 5 channels: 2 for the spatial \textit{xy} offset and 3 for depth, \textit{z}. The final depth is obtained by averaging the 3 depth channels. Opacity ($\alpha$) is represented by a single channel. Covariance ($\Sigma$) is derived from 3 channels of scale and 4 channels of rotation. Finally, color ($c$) is represented using 3 channels. Higher-order spherical harmonics coefficients are not used in our method. The post-activation functions for each parameter follow the design of GS-LRM~\cite{zhang2025gs-lrm}, except for the spatial \textit{xy} offset, for which we constrain the range to lie within a single pixel of viewpoint resolution. We utilize gsplat~\cite{ye2025gsplat}, an open-source library for Gaussian Splatting~\cite{kerbl20233dgs} for a rasterizer. In the post-prediction optimization, we use a learning rate of $6e^{-4}$ for the positions and $1e^{-3}$ for the other attributes.

\noindent\textbf{Camera pose normalization.} We normalize camera poses to align the scene into a consistent coordinate system and scale. First, we compute the average position and viewing directions (forward, down, and right) from the input camera extrinsics. These are used to build a new reference pose, which centers and aligns the scene. All camera extrinsics are then transformed into this reference frame. Finally, we scale the entire scene so that the largest camera distance is 1, ensuring the scene fits within a normalized space~\cite{zhang2025gs-lrm}.

\section{Additional Architectural Details} 
\begin{figure}[!h]
    \centering
    \includegraphics[width=0.7\linewidth]{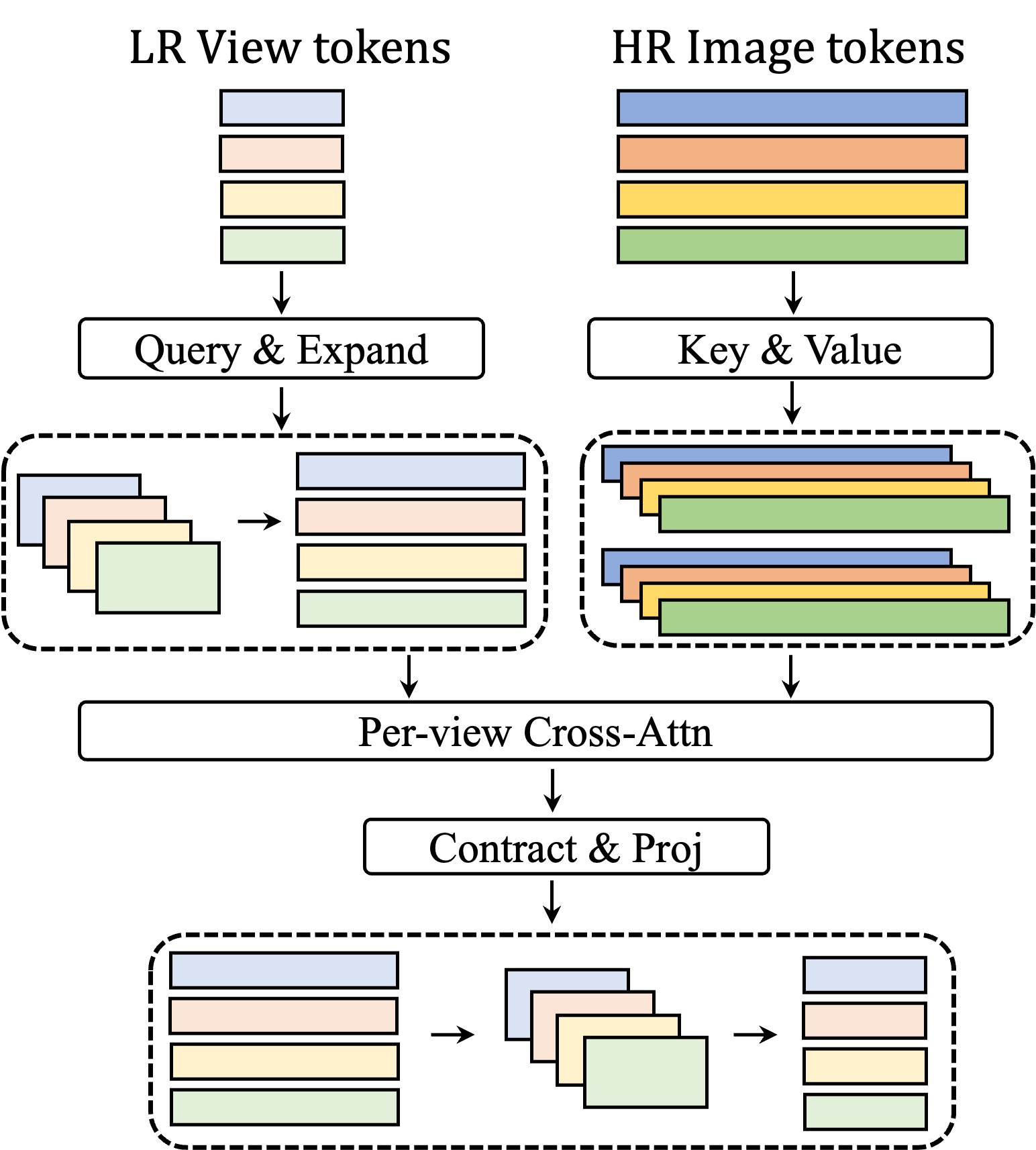}
    \vspace{-2mm}
    \caption{Architectural details of token uplifting.}
    \label{fig:token_uplifting}
    \vspace{-2mm}
\end{figure}

\vspace{-2mm}
We provide the detailed figure of our token uplifting module in Fig.~\ref{fig:token_uplifting}. Note that, to balance the model’s representational capacity and computational efficiency, the length of the low-resolution viewpoint embeddings does not exceed that of the high-resolution image features.

\noindent\textbf{Tokenization and normalization.} After tokenizing the viewpoints and multi-view images using linear layers, both types of tokens are passed through a LayerNorm~\cite{ba2016layernorm}. In each cross-attention layer, only the viewpoint tokens are further processed with a pre-normalization layer. Additionally, after the query and key linear projections, both tokens are passed through an extra normalization layer, referred to as the QK-Norm~\cite{henry2020qknorm}.

\section{Additional Evaluation Details} 
When we utilize more input viewpoints (more than two in RealEstate10K~\cite{zhou2018re10k} experiment compared to the baselines), we sample additional viewpoints/images evenly between the two endpoint indices, ensuring that these samples do not overlap with the target indices. For cross-dataset generalization on the DL3DV dataset, we use a baseline of 12 frames.

In wide-baseline setting, every 8th image in the sequence is reserved for the test split, while K-means clustering on camera positions and viewing directions is applied to the remaining images to select input views that ensure wide scene coverage~\cite{ziwen2025llrm}. 

\begin{table}[!h]
        \centering
        \resizebox{0.9\columnwidth}{!}{%
        \begin{tabular}{r|ccc} 
        \toprule
        & PSNR \(\uparrow\) & SSIM \(\uparrow\)  & LPIPS \(\downarrow\)  \\ 
        \midrule
        Baseline & \textbf{29.24} & \textbf{0.907} & \textbf{0.109} \\
        w/o self-attention & 23.33 & 0.755 & 0.220  \\
        w/ group-attention & 29.02 & 0.904 & 0.112  \\  
        w/ random init. & 28.90 & 0.902 & 0.112  \\          
        w/ LR-feature init. & 28.35 & 0.894 & 0.121  \\  
        \bottomrule
        \end{tabular}
        }
        \caption{
        Additional ablations on model architecture.
        }                
        \label{table:appendix_ablation}
\end{table}

\section{Additional Ablations on Model Architecture} We provide additional ablation studies and analyses in Tab.~\ref{table:appendix_ablation} under the same configuration as the ablations on model architecture in main script. All variants are trained under half-resolution 4 viewpoints setting $(4, H, F)$, with a batch size of 16 on a single RTX 4090 GPU.

\noindent\textbf{1) Self-attention.} To ensure a fair comparison, we replaced all self-attention layers with cross-attention layers rather than simply removing them, maintaining a comparable parameter count. The performance dropped significantly, highlighting the essential role of self-attention in capturing global dependencies and enhancing multi-view awareness among viewpoint embeddings. Without self-attention, the model struggles to integrate contextual information across different viewpoints, resulting in poor convergence and reconstruction quality.

\noindent\textbf{2) Group-attention} This variant replaces the per-viewpoint cross-attention mechanism with a group-attention approach, where all viewpoint tokens and image tokens are concatenated and jointly processed through a cross-attention block. Unlike our default design, group-attention introduces global interactions across all views. While this mechanism can increase the expressive capacity between multiple viewpoints, it incurs quadratic complexity with respect to the number of views. However, the increased computational cost does not yield performance gains, suggesting that separating the roles—using cross-attention for localized image-view interactions and self-attention for global refinement across viewpoints—leads to a more efficient and effective architecture, which is also validated as Alternating-Attention in VGGT~\cite{wang2025vggt}.

\noindent\textbf{3) Different initialization.} We also investigate the different initialization methods of scene representation. For the random initizliation, we used a learnable embedding initialized with zero mean and 0.02 standard deviation, whereas for the LR feature variant, we used features extracted from low-resolution images. In the PSNR training curve, the LR feature variant rises more quickly in the early stages but is later surpassed by the random initialization variant. We believe this is because, in our iterative cross-attention architecture, high-resolution image features and camera information are continually provided by the cross-attention blocks. As a result, the learnable embedding can offer more flexible and informative parameters for guiding iterative updates, whereas the LR image features may introduce redundant and less discriminative information that limits long-term performance gains. Moreover, the use of LR features may bias the early attention stages toward coarse processing, which can hinder the model’s ability to fully refine fine details in later stages.

\section{Computational Costs of Training} 
\vspace{-2mm}
We provide a detailed theoretical calculation of the FLOPs for our mini-batch cross-attention mechanism.  In this analysis, we limit the computation to a per-view, single cross-attention operation, excluding our token uplifting strategy (as it introduces a constant cost across all variations). Given a viewpoint token of shape $(L_v, D)$ and an image token of shape $(L_i, D)$, where $L_v$ and $L_i$ denote the token lengths and $D$ is the hidden dimension, the FLOPs for the cross-attention operation are computed as:
\[
4D^2(L_v + L_i) + 4L_v L_i D.
\]
Assuming a hidden dimension of $D = 768$, an image resolution of $256 \times 256$, and a viewpoint resolution of $128 \times 128$, with a patch size of $8 \times 8$, the token lengths are computed as $L_i = 1024$ for the image tokens and $L_v = 256$ for the viewpoint tokens, based on the experimental configuration used in the RealEstate10K~\cite{zhou2018re10k} dataset.

Thus, the computation becomes: baseline: \textbf{3.83 GFLOPs}; half cross-attention: \textbf{1.71 GFLOPs}; quarter cross-attention: \textbf{0.81 GFLOPs}.

\section{Additional Quantitative Results}

\noindent\textbf{Wide-coverage baselines.} We additionally evaluate our method on the recently released DL3DV~\cite{ling2024dl3dv} evaluation split, which comprises 51 scenes in our experiments. Our method achieves better reconstruction quality, faster inference, and stronger longer-context generalization, while its compact 3D scene representations additionally enable fast rendering, as shown in Tab.~\ref{tab:undisthr_table_supple}.

\begin{table}[!h]
    \centering
    \resizebox{1.0\columnwidth}{!}{
        \begin{tabular}{lccccc}
        \toprule
        Method & Views & Time $\downarrow$ & PSNR $\uparrow$ & SSIM $\uparrow$ & LPIPS $\downarrow$  \\
        \midrule
        3D-GS~\cite{kerbl20233dgs} & 32 & 8min & 25.09 & 0.838 & 0.175  \\     
        \midrule
        Long-LRM~\cite{ziwen2025llrm} & 32 & 0.84sec & 23.54 & 0.776 & 0.270  \\
        Ours & $(32, H, F)$ & \best{0.53sec} & \best{23.93} & \best{0.800} & \best{0.259}  \\
        \midrule
        \midrule
        Long-LRM (Unseen) & 16 & 0.50sec & 20.65 & 0.707 & 0.328  \\
        Ours (Unseen) & $(16, H, F)$ & \best{0.19sec} & \best{21.63} & \best{0.746} & \best{0.316}  \\           
        \midrule
        Long-LRM (Unseen) & 40 & 1.05sec & 23.76 & 0.785 & 0.262  \\
        Ours (Unseen) & $(40, H, F)$ & \best{0.76sec} & \best{24.21} & \best{0.809} & \best{0.250}  \\   
        \midrule
        Long-LRM (Unseen) & 48 & 1.38sec & 23.88 & 0.795 & 0.255  \\
        Ours (Unseen) & $(48, H, F)$ & \best{1.04sec} & \best{24.45} & \best{0.818} & \best{0.242}  \\  
        \bottomrule
        \end{tabular}
    }
    \vspace{-2mm}    
    \caption{
    Quantitative comparisons on the undistorted DL3DV evaluation dataset (540$\times$960). We utilized flash attention v3~\cite{shah2024flashattention} using a H100 GPU. 
    } 
    \label{tab:undisthr_table_supple}   
    \vspace{-5mm}
\end{table}

\begin{figure}[!h]
    \centering
    \includegraphics[width=1.0\linewidth]{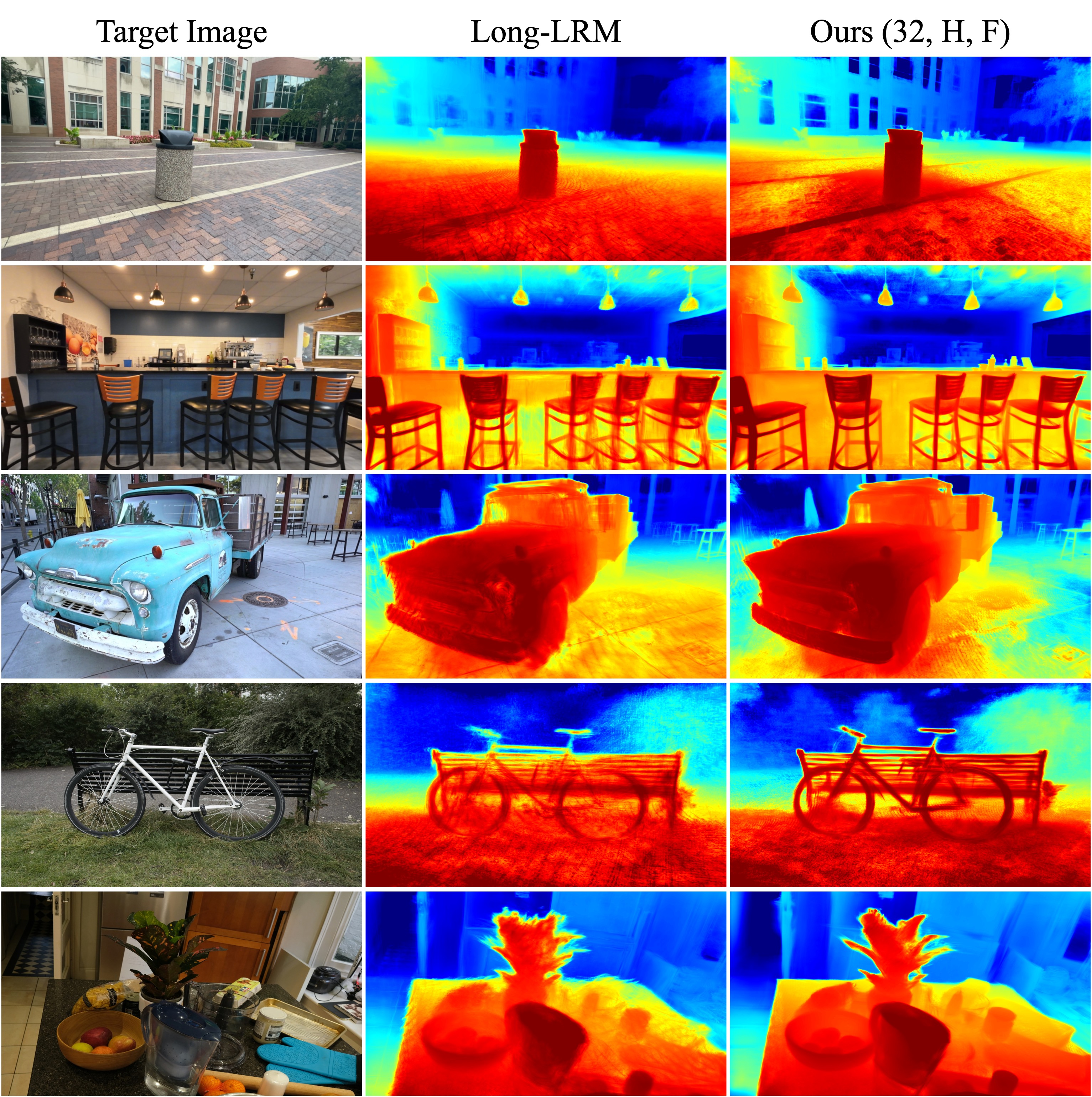}    
    \vspace{-5mm}    
    \caption{Qualitative comparison of rendered depth maps. Examples from DL3DV (top two rows), Tanks\&Temples (third row), and Mip-NeRF360 (bottom two rows) are shown.} 
    \label{fig:qual_depth}
\end{figure}

\noindent\textbf{Depth estimation.} In addition to novel view synthesis, we further evaluate the rendered depth maps, which serve as a indicator measure for underlying geometric accuracy, on the DL3DV~\cite{ling2024dl3dv}, Tanks\&Temples (TNT)~\cite{Knapitsch2017tnt}, and Mip-NeRF360 (Mip360)~\cite{barron2022mip} dataset. Since these datasets do not provide ground-truth depth, we adopt the recent state-of-the-art monocular depth estimator, MoGe-2~\cite{wang2025moge}, as a proxy to obtain pseudo depth. For each target view, we predict depth from the target image and its focal length, mask out invalid values in both rendered and pseudo depths, and then compute relative depth accuracy by comparing them using standard scale-invariant depth metrics. Tab.~\ref{tab:depth_estimation} shows that our method produces depth maps that are more consistent with the MoGe-2 predictions than those of the baseline, even though the baseline is additionally regularized using a pretrained depth estimation model~\cite{depthanything} during training. We regard this evaluation as an indirect indicator of geometric quality in the absence of ground-truth depth. We also provide qualitative visualizations of the rendered depth maps in Fig.~\ref{fig:qual_depth}, where our method produces sharper, more detailed depth boundaries and fewer artifacts compared to the baseline. We attribute these improvements to our compact representation, which reduces redundancy and artifacts while better preserving fine geometric details.

\begin{table}[!h]
    \centering
    \resizebox{1.0\columnwidth}{!}{
        \begin{tabular}{@{}lccccccc@{}}
        \toprule
        \multirow{2}{*}{Method} & Views & \multicolumn{2}{c}{DL3DV-Benchmark} & \multicolumn{2}{c}{DL3DV-Eval} & \multicolumn{2}{c}{TNT\&Mip360} \\
        \cmidrule(lr){3-4}\cmidrule(lr){5-6} \cmidrule(lr){7-8}     
        & & Abs Rel $\downarrow$ & RMSE $\downarrow$ & Abs Rel $\downarrow$ & RMSE $\downarrow$ & Abs Rel $\downarrow$ & RMSE $\downarrow$ \\
        \midrule
        Long-LRM~\cite{ziwen2025llrm}  & 16 & 0.718  & 1.427 & 0.768 & 1.631 & 0.603 & 1.199 \\
        Ours & $(16, H, F)$ & \best{0.670} & \best{1.174} & \best{0.693} & \best{1.354} & \best{0.515} & \best{0.852} \\
        \midrule
        Long-LRM  & 32 & 0.759 & 1.571 & 0.805 & 1.788 & 0.645 & 1.345 \\
        Ours & $(32, H, F)$ & \best{0.709} & \best{1.310} & \best{0.739} & \best{1.513} & \best{0.543} & \best{0.967} \\ 
        \bottomrule
        \end{tabular}        
    }
    \vspace{-2mm}    
    \caption{Quantitative comparison of depth estimation. We use the large MoGe-2 (ViT-L) variant as the pseudo-depth estimator.}
    \label{tab:depth_estimation}
    \vspace{-2mm}
\end{table}

\begin{figure}[!h]
    \centering
    \vspace{-2mm}
    \includegraphics[width=1.0\linewidth]{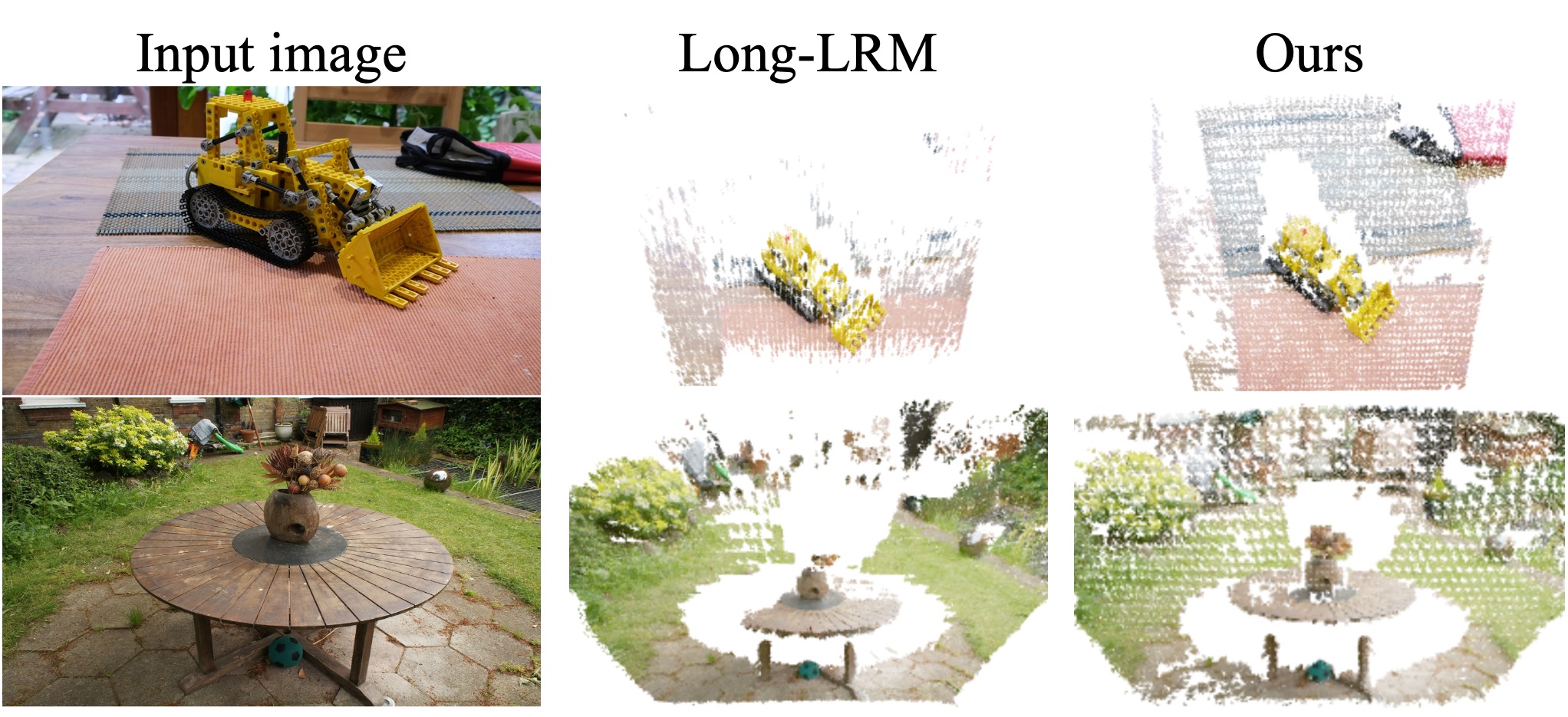}
    \vspace{-3mm}
    \caption{Zero-shot colored vertices visualization.}
    \label{fig:pointmap}
\end{figure}

\noindent\textbf{Geometry estimation} We evaluate the Chamfer Distance (CD) and F1-score with geometry datasets using NRGBD~\cite{azinovic2022nrgbd} and ETH3D~\cite{eth}. For iLRM, ground-truth pointmaps are downsampled by half to match the generated point clouds. With Mip-NeRF360 dataset, we visualized the vertices after building mesh and removing flying points (relative tolerance threshold of 0.04). iLRM shows better geometry with fewer points in Tab.~\ref{tab:pointmap} and Fig.~\ref{fig:pointmap}. 

\begin{table}[!h]
    \renewcommand{\arraystretch}{0.85}
    \centering
    \resizebox{1.0\columnwidth}{!}{
        \begin{tabular}{lcccccccc}
        \toprule
        \multirow{2}{*}{Method} & \multicolumn{2}{c}{NRGBD (16-view)} &
        \multicolumn{2}{c}{NRGBD (32-view)} & \multicolumn{2}{c}{ETH3D (16-view)} &
        \multicolumn{2}{c}{ETH3D (32-view)} \\
        \cmidrule(lr){2-3}\cmidrule(lr){4-5} \cmidrule(lr){6-7}\cmidrule(lr){8-9}     & CD $\downarrow$ & F1-score $\uparrow$ & CD $\downarrow$ & F1-score $\uparrow$ & CD $\downarrow$ & F1-score $\uparrow$ & CD $\downarrow$ & F1-score $\uparrow$ \\
        \midrule
        Long-LRM   & 0.53 & 0.52 & \best{0.43} & 0.59 & 2.75 & 0.32 & 2.69 & 0.39 \\
        Ours & \best{0.50} & \best{0.60} & 0.52 & \best{0.69} & \best{2.06} & \best{0.50} & \best{1.15} & \best{0.54} \\      
        \bottomrule
        \end{tabular}        
    }
    \vspace{-3mm}
    \caption{Pointmap estimation comparisons with a input resolution of 540 × 960 (\# views). The F1-score threshold was set to 0.1.}
    \label{tab:pointmap}
\end{table}

\begin{figure}[!h]
    \centering
    \includegraphics[width=1.0\linewidth]{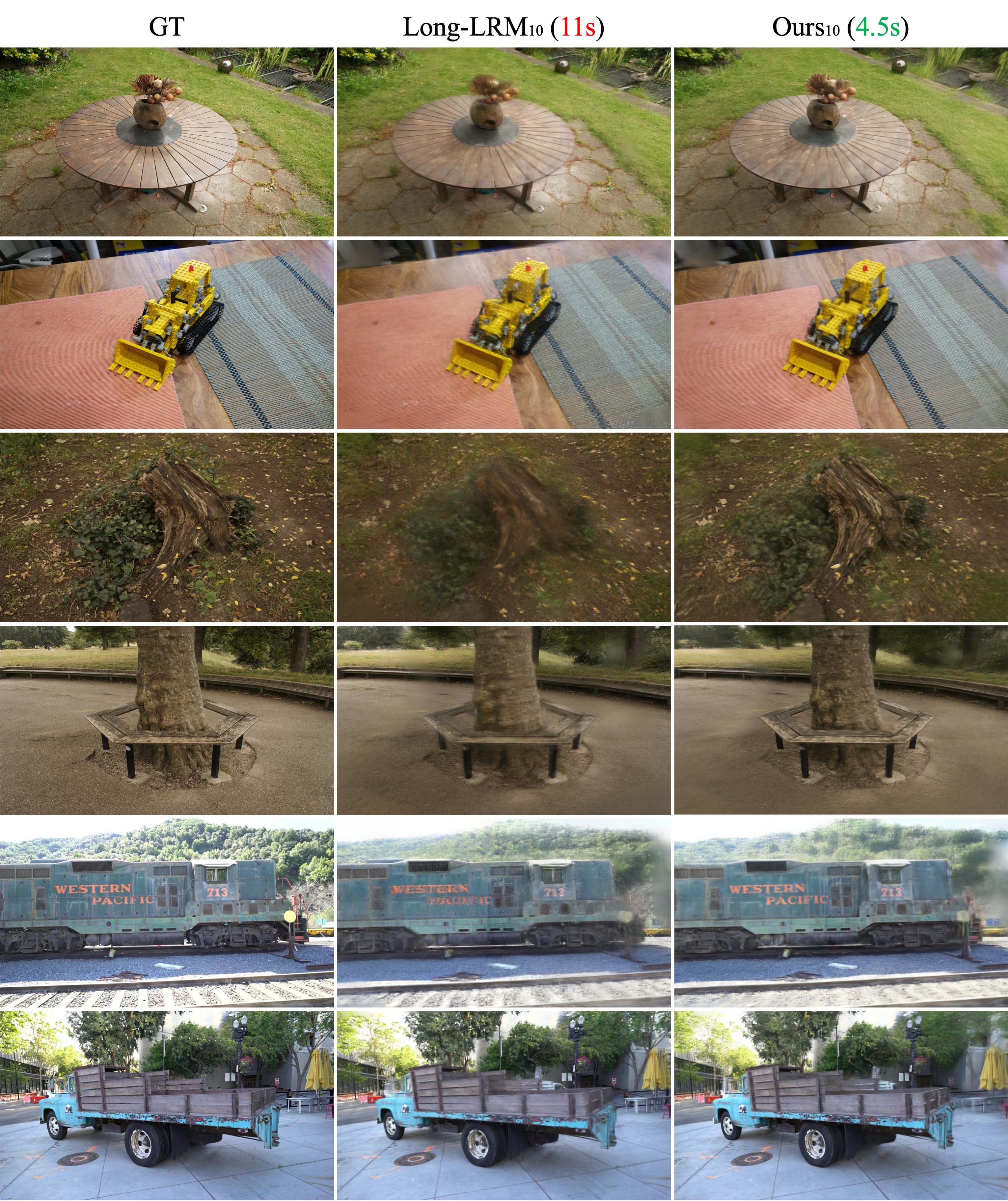}    
    \vspace{-6mm}    
    \caption{Qualitative comparison on Mip-NeRF360 (first four rows) and Tanks\&Temples (bottom two rows) after 10-epoch of post-prediction optimization.} 
    \label{fig:ft10}
    \vspace{-1mm}    
\end{figure}

\noindent\textbf{Post-prediction optimization} While the performance of zero-shot novel view synthesis (TNT, Mip360) in Tab.~\ref{tab:hrcross} is comparable to the baseline, our finer geometric estimation promote faster convergence during post-prediction optimization which reflects more reliable geometry in the initial 3D Gaussians. In 10-epoch optimization, our method already outperforms the baseline while using less than half of the time, and with 20-epoch, still faster than the baseline, it achieves even higher accuracy. We attribute these gains to our compact scene representation and its stronger capacity for capturing underlying geometric structure. We provide qualitative comparisons on 10-epoch optimization in Fig.~\ref{fig:ft10}.

\begin{table}[!h]
    \centering
    \resizebox{1.0\columnwidth}{!}{
        \renewcommand{\arraystretch}{1.0}
        \begin{tabular}{lccccccccc}
        \toprule
        \multirow{2}{*}{Method} & \multirow{2}{*}{Views} & \multirow{2}{*}{Time $\downarrow$} & \multicolumn{3}{c}{Tanks\&Temples~\cite{Knapitsch2017tnt}} & \multicolumn{3}{c}{Mip-NeRF360~\cite{barron2022mip}} \\
        \cmidrule(lr){4-6}\cmidrule(lr){7-9}
        & & & PSNR $\uparrow$ & SSIM $\uparrow$ & LPIPS $\downarrow$ & PSNR $\uparrow$ & SSIM $\uparrow$ & LPIPS $\downarrow$ \\  
        \midrule
        3D-GS~\cite{kerbl20233dgs} & 32 & 8min & 18.48 & 0.720 & 0.260 & 22.95 & 0.694 & 0.250 \\       
        \midrule
        Long-LRM~\cite{ziwen2025llrm} & 32 & 0.84sec & \best{18.59} & 0.614 & \best{0.367} & 21.08 & 0.484 & \best{0.445} \\
        Ours & $(32, H, F)$ & \best{0.53sec} & 18.58 & \best{0.631} & 0.385 & \best{21.09} & \best{0.495} & 0.466 \\
        \midrule
        Long-LRM$_{10}$ & 32 & 11sec & 19.23 & 0.663 & 0.348 & 22.05 & 0.554 & 0.414 \\
        Ours$_{10}$ & $(32, H, F)$ & \best{4.5sec} & 19.42 & 0.689 & 0.350 & 22.49 & 0.601 & 0.414 \\ 
        Ours$_{20}$ & $(32, H, F)$ & 8.6sec & \best{19.62} & \best{0.704} & \best{0.338} & \best{22.85} & \best{0.622} & \best{0.397} \\         
        \bottomrule
        \end{tabular}
    }
    \vspace{-2mm}    
    \caption{Quantitative comparisons on the Tanks\&Temples and Mip-NeRF360 dataset (540$\times$960). We adopt the settings reported in their paper for the post-prediction optimization of Long-LRM (learning rates of $5e^{-4}$ for position and $1e^{-3}$ for color). We utilized flash attention v3~\cite{shah2024flashattention} using a H100 GPU. }
    \label{tab:hrcross}    
\end{table}

\noindent\textbf{Input robustness.}
To assess the robustness of our model to imperfect camera pose estimates, we evaluate under translational camera pose perturbations on the 32-view DL3DV dataset. Gaussian noise is added to the camera translation vectors, and we report performance across varying noise to examine the degradation in rendering quality (Tab.~\ref{tab:noise}).

\begin{table}[!h]    
    \centering
    \resizebox{1.0\columnwidth}{!}{
        \renewcommand{\arraystretch}{1.0}
        \begin{tabular}{lcccccc}
        \toprule
        \multirow{2}{*}{stand. dev.} & \multicolumn{2}{c}{0} & \multicolumn{2}{c}{0.001} & \multicolumn{2}{c}{0.005} \\
        \cmidrule(lr){2-3}\cmidrule(lr){4-5}\cmidrule(lr){6-7}
        & PSNR & LPIPS & PSNR & LPIPS & PSNR & LPIPS\\
        \midrule
        Long-LRM & 23.97 & 0.267 & 23.27 (-0.70) & 0.287 (+0.020) & 20.49 (-3.48) &0.381 (+0.114) \\     
        iLRM & \best{24.30} & \best{0.256} & \best{23.82 (-0.48)} & \best{0.270 (+0.014)} & \best{21.49 (-2.81)} & \best{0.352 (+0.096)} \\                  
        \bottomrule
        \end{tabular}        
    }
    \caption{Robustness evaluation under translational camera pose perturbations on the 32-view DL3DV (540$\times$960) dataset.}
    \label{tab:noise}
\end{table}

\noindent\textbf{Inference time.} We compare the inference time of our model across different numbers of input views at a resolution of 540$\times$960, using a single H100 GPU with flash attention v3~\cite{shah2024flashattention}. As shown in Tab.~\ref{tab:inference}, our method achieves lower latency than Long-LRM across all comparable settings. Notably, Long-LRM runs out of memory at 256 input views, whereas our model still completes inference within a practical time budget, indicating better scalability of the proposed compact viewpoint representation to large number of views.

\begin{table}[!h]
    \centering
    \resizebox{1.0\columnwidth}{!}{
        \renewcommand{\arraystretch}{1.0}
        \begin{tabular}{@{}lcccccc@{}}
        \toprule
        Method & 16 & 32 & 64 & 96 & 128 & 256 \\
        \midrule
        Long-LRM~\cite{ziwen2025llrm} & 0.5 & 0.84 & 2.08 & 3.90 & 6.39 & Out-of-memory \\
        Ours & \best{0.19} & \best{0.53} & \best{1.66} & \best{3.37} & \best{5.61} & \best{20.92} \\
        \bottomrule
        \end{tabular}
    }    
    \caption{Quantitative comparisons of inference time across different numbers of input views. All times are measured in seconds.}
    \label{tab:inference}   
\end{table}

\begin{table*}[!h]
    \centering
    \resizebox{1.0\linewidth}{!}{
    \begin{tabular}{lcccccccccccc}
            \toprule
               & \multicolumn{3}{c}{Small}                                                       & \multicolumn{3}{c}{Medium}                                                      & \multicolumn{3}{c}{Large}                                                       & \multicolumn{3}{c}{Average}                                                     \\ \cmidrule(lr){2-4} \cmidrule(lr){5-7} \cmidrule(lr){8-10} \cmidrule(lr){11-13} 
    Method     & \multicolumn{1}{c}{PSNR $\uparrow$} & \multicolumn{1}{c}{SSIM $\uparrow$} & \multicolumn{1}{c}{LPIPS $\downarrow$} & \multicolumn{1}{c}{PSNR $\uparrow$} & \multicolumn{1}{c}{SSIM $\uparrow$} & \multicolumn{1}{c}{LPIPS $\downarrow$} & \multicolumn{1}{c}{PSNR $\uparrow$} & \multicolumn{1}{c}{SSIM $\uparrow$} & \multicolumn{1}{c}{LPIPS $\downarrow$} & \multicolumn{1}{c}{PSNR $\uparrow$} & \multicolumn{1}{c}{SSIM $\uparrow$} & \multicolumn{1}{c}
    {LPIPS $\downarrow$} \\
    \midrule
    MVSplat~\cite{chen2025mvsplat}   & 20.37 & 0.725 & 0.250 & 23.81 & 0.814 & 0.172 & 27.47 & 0.885 & 0.115 & 24.01 & 0.812 & 0.175 \\
    DepthSplat~\cite{xu2025depthsplat} & 22.82 & 0.798 & 0.193 & 25.38 & 0.851 & 0.145 & 28.32 & 0.900 & 0.104 & 25.59 & 0.852 & 0.145 \\
    Gen-Den~\cite{nam2025generative} & 21.10 & 0.744 & 0.234 & 24.57 & 0.828 & 0.162 & 28.26 & 0.895 & 0.108 & 24.77 & 0.826 & 0.164 \\
    Ours $(2, F, F)$ & \best{23.82} & \best{0.813} & \best{0.184} & \best{26.54} & \best{0.864} & \best{0.139} & \best{29.43} & \best{0.910} & \best{0.103} & \best{26.70} & \best{0.864} & \best{0.140} \\ 
    \midrule
    Ours $(4, H, F)$ & 27.65 & 0.887 & 0.127 & 29.13 & 0.908 & 0.108 & 30.73 & 0.926 & 0.092 & 29.22 & 0.908 & 0.108 \\
    Ours-MC $(4, H, F)$ & 27.41 & 0.882 & 0.131 & 28.87 & 0.904 & 0.111 & 30.44 & 0.927 & 0.095 & 28.96 & 0.904 & 0.112 \\    
    Ours $(8, H, F)$ & 29.44 & 0.912 & 0.106 & 30.51 & 0.925 & 0.093 & 31.77 & 0.937 & 0.080 & 30.61 & 0.925 & 0.092 \\
    Ours-MC $(8, H, F)$ & 29.15 & 0.908 & 0.108 & 30.20 & 0.922 & 0.094 & 31.46 & 0.935 & 0.082 & 30.30 & 0.922 & 0.094 \\   \bottomrule         
    \end{tabular}}
    \vspace{-2mm}      
    \caption{
    Quantitative comparisons on the RE10K dataset under varying view overlap conditions.
    }        
    \label{tab:nopo quantitative result on re10k}
    \vspace{-2mm}     
\end{table*}

\begin{table*}[!h]
    \centering
    \resizebox{0.9\linewidth}{!}{
        \begin{tabular}{lcccccccc}
        \toprule
        Method & Params (M) & Train GPU (\#) & PSNR $\uparrow$ & SSIM $\uparrow$ & LPIPS $\downarrow$ & \# Gaussians & Time (s) & Memory (GB) \\
        \midrule
        MVSplat~\cite{chen2025mvsplat} & 12 & H100 (1) & 27.53 & 0.889 & 0.116 & 65,536 & 0.048 & \best{0.65} \\
        DepthSplat~\cite{xu2025depthsplat} & 354 & H100 (1) & 28.08 & 0.898 & 0.107 & 65,536 & 0.062 & 2.49 \\
        Ours & 185 & RTX 4090 (1) & 29.24 & 0.907 & 0.109 & 65,536 & \best{0.027} & 1.22 \\
        Ours & 185 & RTX 4090 (2) & \best{29.82} & \best{0.916} & \best{0.101} & 65,536 & \best{0.027} & 1.22 \\
        \bottomrule
        \end{tabular}
    }
    \vspace{-2mm}    
    \caption{
    Quantitative comparisons under the same number of Gaussians on the RE10K dataset. Inference time and memory consumption are measured only during the Gaussian generation stage, excluding the rendering process on a RTX 4090 GPU.
    }          
    \label{tab:lr_baseline}  
    \vspace{-3mm}
\end{table*}

\noindent\textbf{Varying baseline range.} We compare our model against recent generalizable 3D reconstruction methods~\cite{chen2025mvsplat, xu2025depthsplat, nam2025generative} on the RealEstate10K~\cite{zhou2018re10k} dataset, with a particular focus on handling varying degrees of camera overlap~\cite{ye2024no}. These overlap categories are determined using the dense feature matching method RoMA~\cite{edstedt2024roma}. As shown in Tab.~\ref{tab:nopo quantitative result on re10k}, our method, which efficiently handles a large number of input viewpoints/images, achieves superior performance compared to existing approaches, especially in challenging cases with small viewpoint overlap.

\noindent\textbf{Same number of Gaussians.} 
We also validate the strength of our decoupling strategy in leveraging high-resolution images as visual cues while generating efficient and compact 3D Gaussians. As discussed in our motivation, previous methods~\cite{charatan2024pixelsplat, chen2025mvsplat, zhang2025gs-lrm, xu2025depthsplat, nam2025generative} require downsampling the input images to reduce the number of generated Gaussians, inherently coupling image resolution with representation density. To demonstrate the flexibility of our approach, we conduct an experiment in which all methods generate the same number of Gaussians using 4 viewpoints at half resolution. Specifically, the baseline methods~\cite{chen2025mvsplat, xu2025depthsplat} follow a $(4, H, H)$ configuration, where both the number of viewpoints and image resolution are reduced. In contrast, our method adopts a $(4, H, F)$ setting, where we preserve high-resolution image inputs while generating low-resolution Gaussians, thanks to our decoupled design. As shown in Tab.~\ref{tab:lr_baseline}, our method surpasses the baselines in performance while requiring fewer computational resources in training, and faster inference speed, highlighting the practical advantages of our design. This result demonstrates the efficiency and the representational ability of our architecture, which effectively utilizes high-resolution visual cues, leading to superior reconstruction quality under the same output density without requiring expensive hardware. To train the baseline methods effectively with a large batch size (similar to ours), we run them on a single H100 GPU. Our method and MVSplat~\cite{chen2025mvsplat} are trained with a batch size of 16, while DepthSplat~\cite{xu2025depthsplat} is trained with a batch size of 12 due to memory constraints.

\noindent\textbf{Quarter resolution baselines.} To evaluate different viewpoint configurations—specifically the resolution of each viewpoint—we additionally compare a quarter-resolution variant of the viewpoint inputs. All experiments in Tab.~\ref{tab:viewpoint resolution on re10k} are conducted using a single RTX 4090 GPU with a batch size of 16, 12 update layers, and trained for 100,000 iterations. While lowering the resolution of viewpoint inputs leads to a moderate drop in reconstruction quality, it significantly reduces the number of generated Gaussians, showing trade-off between accuracy and efficient representations.

\begin{table}[!h]
    \centering
    \resizebox{1.0\columnwidth}{!}{
        \begin{tabular}{lcccc}
        \toprule
        Method & PSNR $\uparrow$ & SSIM $\uparrow$ & LPIPS $\downarrow$ & \# Gaussians \\
        \midrule
        Ours $(8, H, F)$ & \best{30.39} & \best{0.923} & \best{0.095} & 131,072 \\
        Ours $(4, H, F)$ & 29.24 & 0.907 & 0.109 & 65,536 \\      
        Ours $(8, Q, F)$ & 27.36 & 0.868 & 0.152 & 32,768\\  
        Ours $(4, Q, F)$ & 26.40 & 0.843 & 0.177 & \best{16,384}\\  
        \bottomrule
        \end{tabular}
    }
    \vspace{-2mm}
    \caption{
Quantitative comparisons of different viewpoint configurations on the RE10K dataset. $Q$ denotes quarter resolution compared to the original image resolution.
    }      
    \label{tab:viewpoint resolution on re10k}   
\end{table}

\section{Additional Qualitative Results}
\begin{figure*}[!h]
    \centering
    \includegraphics[width=1.0\linewidth]{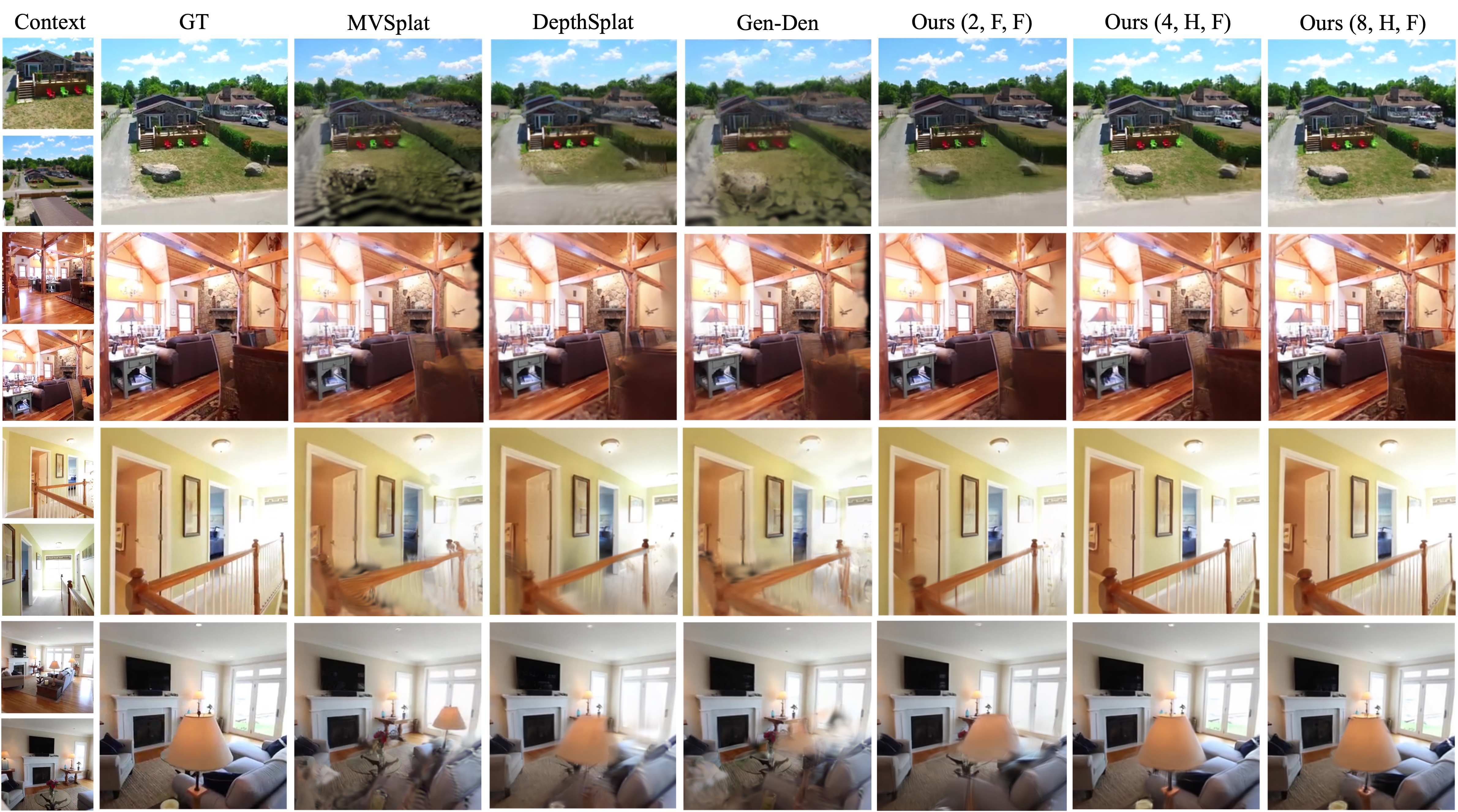}
    \caption{Qualitative comparison on the RE10K dataset (2 input images except for ``Ours(4, H, F)" and ``Ours(8, H, F)", 256$\times$256).}    
    \label{fig:sub_re10k}
\end{figure*}

\begin{figure*}[!h]
    \centering
    \includegraphics[width=1.0\linewidth]{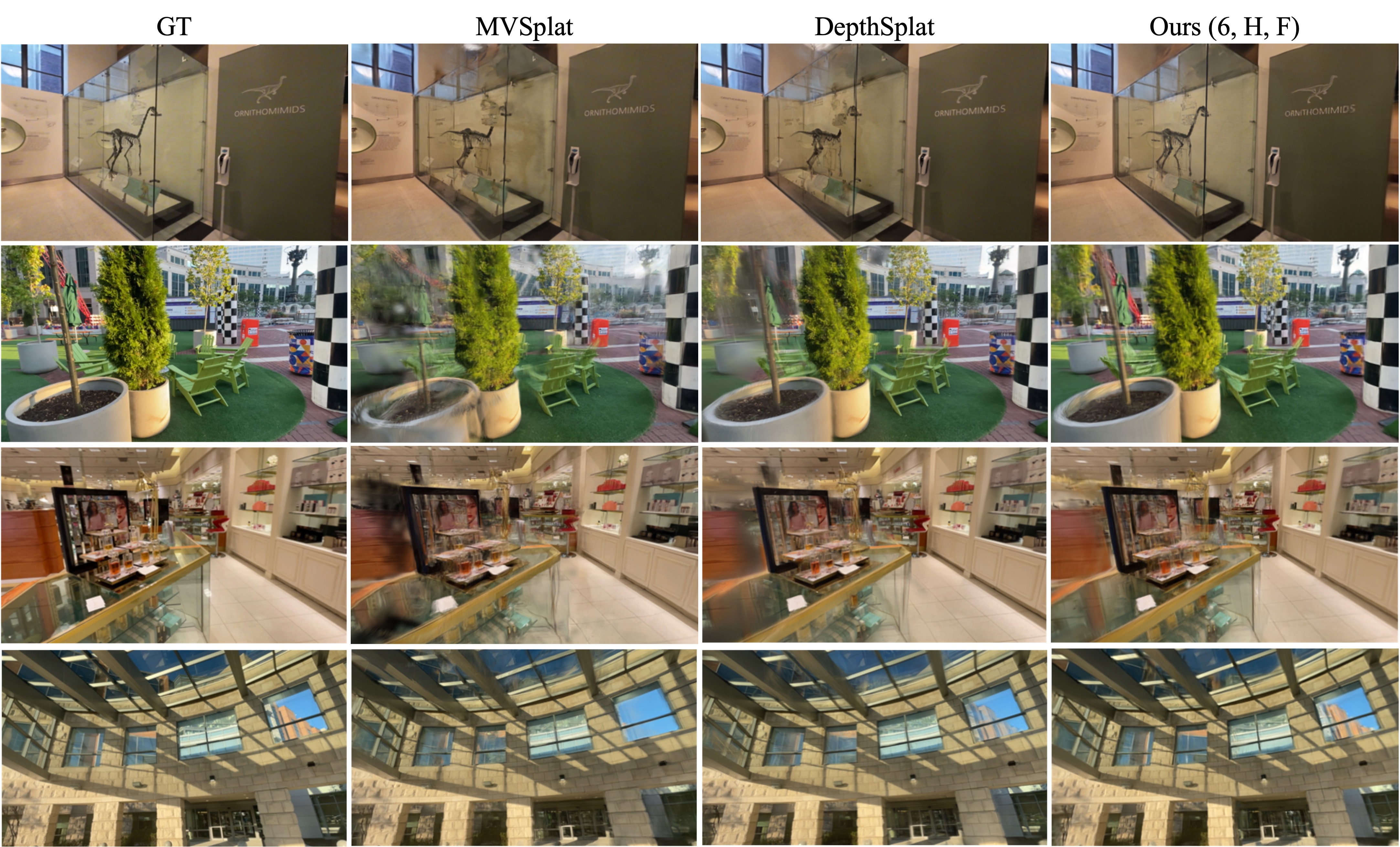}
    \caption{Qualitative comparison on the DL3DV dataset under the 50-frame baseline setting (6 input images, 256$\times$448).}
    \label{fig:sub_dl3dv}
\end{figure*}

\begin{figure*}[!h]
    \centering
    \includegraphics[width=1.0\linewidth]{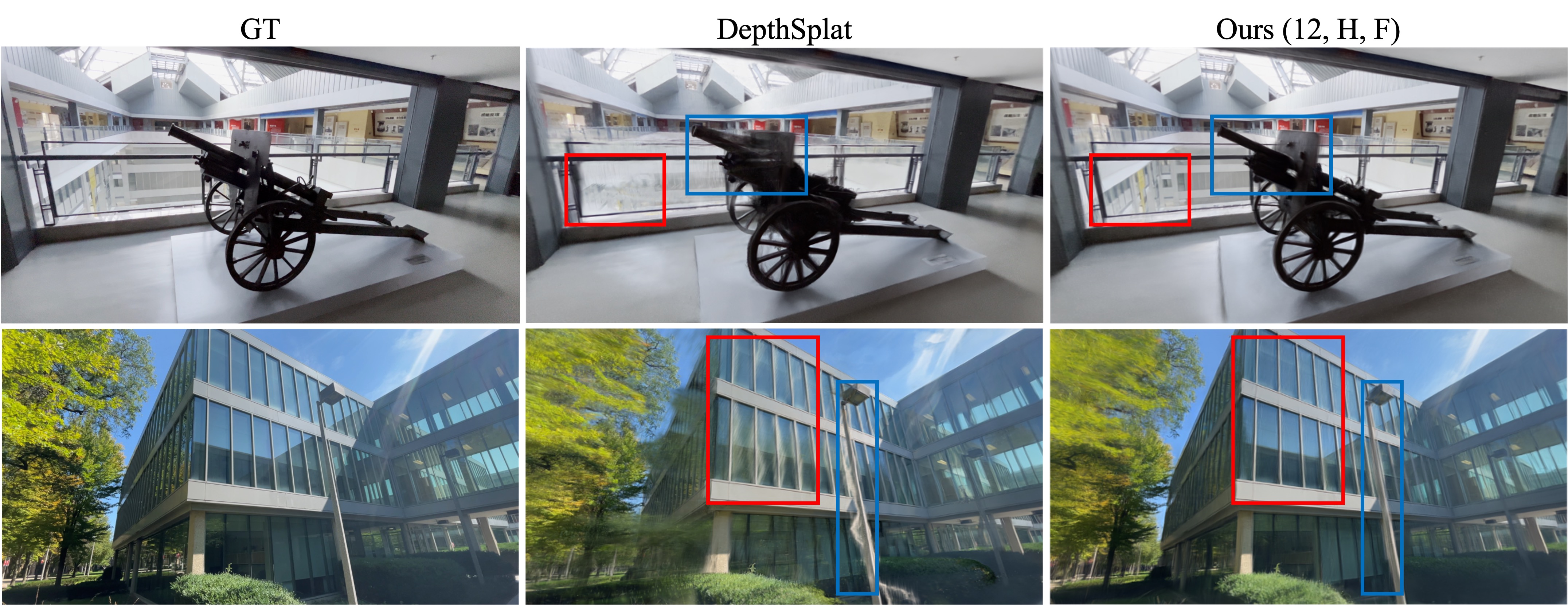}
    \caption{Qualitative comparison on the DL3DV dataset under the 100-frame baseline setting (12 input images, 512$\times$960).}
    \label{fig:sub_dl3dv_hr}
\end{figure*}

\begin{figure*}[!h]
    \centering
    \includegraphics[width=1.0\linewidth]{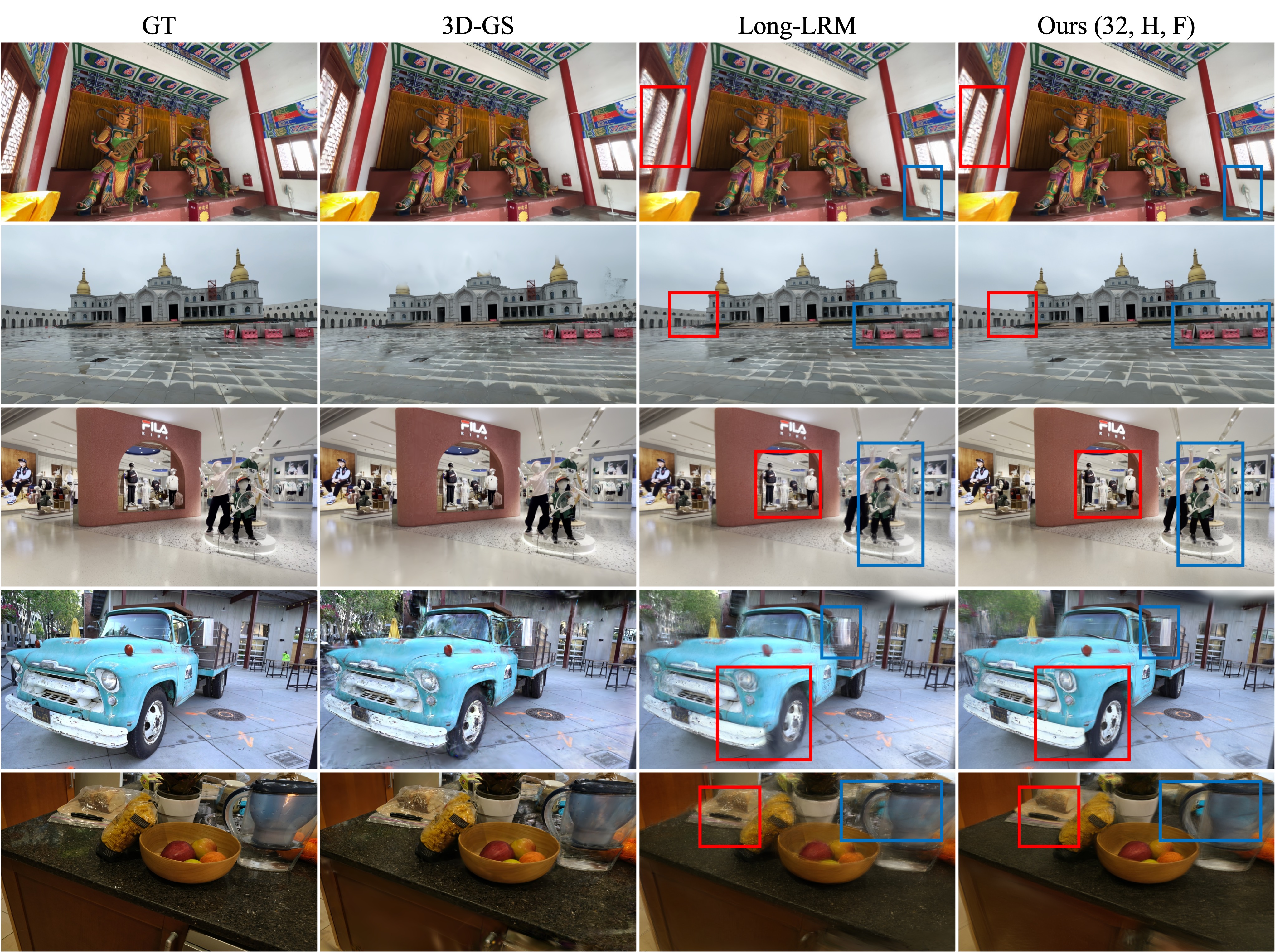}
    \caption{Qualitative comparison under the wide-baseline setting (32 input images, 540$\times$960, zero-shot). DL3DV (top three rows), Tanks\&Temples (fourth row), and Mip-NeRF360 (bottom row) are shown.}
    \label{fig:sub_dl3dv_gs}
\end{figure*}

\begin{figure*}[!h]
    \centering
    \includegraphics[width=1.0\linewidth]{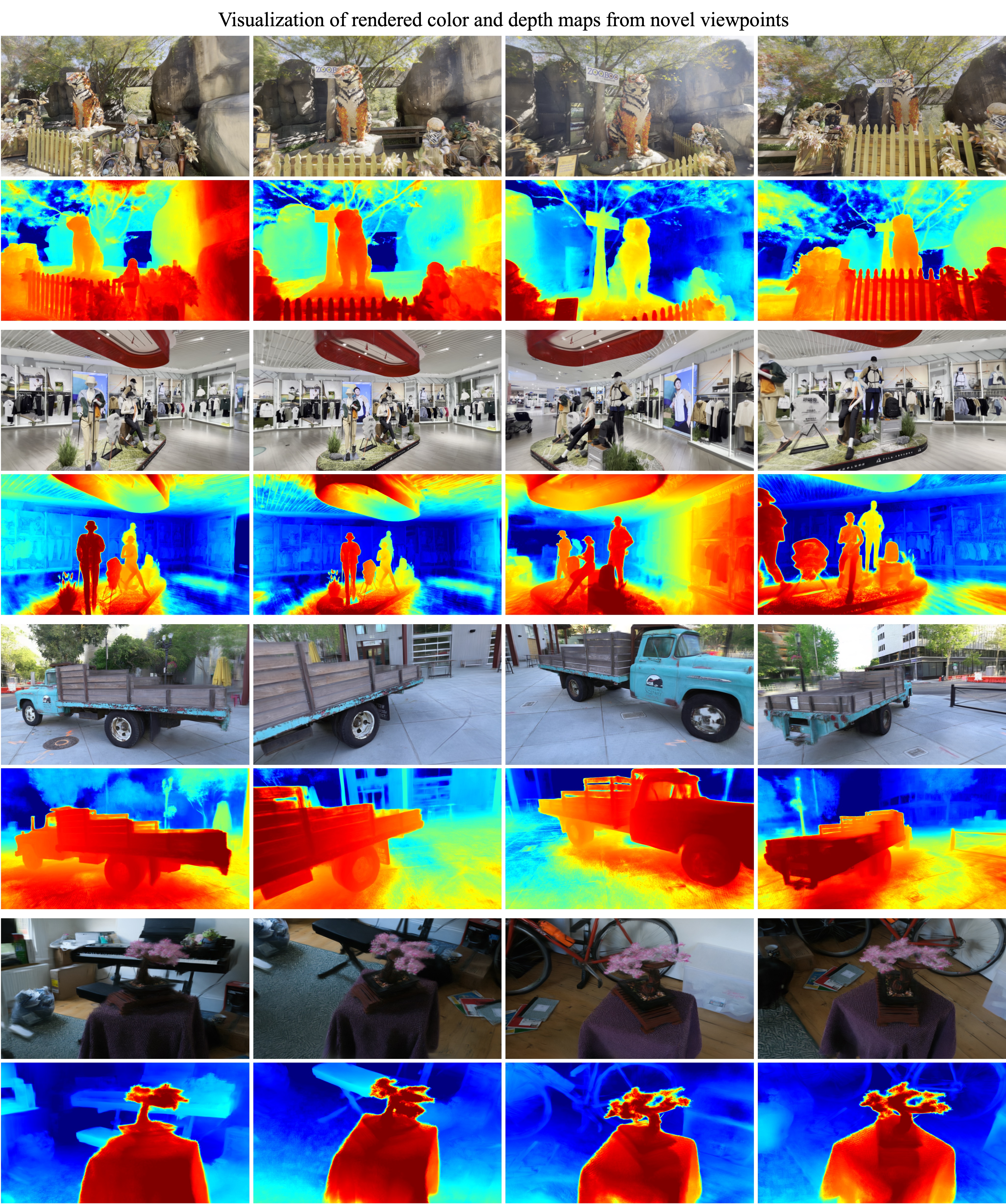}
    \caption{Qualitative visualization of rendered color and depth maps from novel viewpoints (32 input images, 540$\times$960, zero-shot). Scenes from DL3DV (top two example), Tanks\&Temples (third example), and Mip-NeRF360 (bottom example) are shown.}
    \label{fig:rgb_depth}
\end{figure*}

\begin{figure*}[!h]
    \centering
    \includegraphics[width=1.0\linewidth]{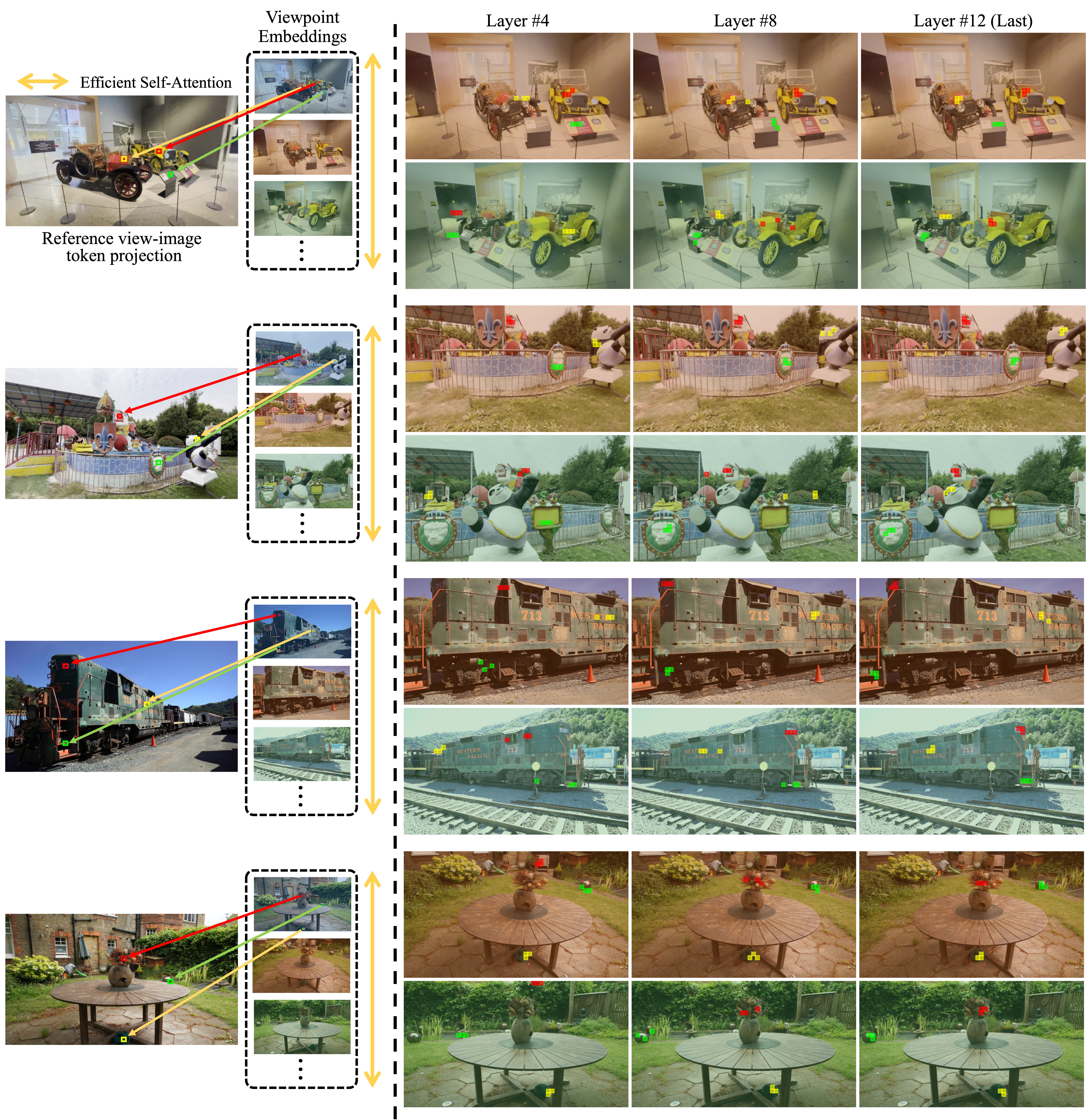}
    \caption{For the colored query patches in the reference viewpoint (\textcolor[HTML]{C02A1E}{red}, \textcolor[HTML]{D6D659}{yellow}, \textcolor[HTML]{62CF42}{green}), we visualize top-3 attended tokens from other viewpoints throughout the iterative refinement process. For relatively easy cases with small camera motion or distinctive regions, the model identifies the correct correspondences in the early layers, whereas for more challenging cases with larger viewpoint changes, the attention gradually converges to geometrically consistent regions as the refinement progresses. Scenes from DL3DV (top two rows), Tanks\&Temples (third row), and Mip-NeRF360 (bottom row) are shown.}
    \label{fig:att_sup}
\end{figure*}

We present additional qualitative results in Fig.~\ref{fig:sub_re10k} for the RealEstate10K (RE10K)~\cite{zhou2018re10k} dataset and in Fig.~\ref{fig:sub_dl3dv},~\ref{fig:sub_dl3dv_hr}, ~\ref{fig:sub_dl3dv_gs} and~\ref{fig:rgb_depth} for the DL3DV~\cite{ling2024dl3dv}, Tanks\&Temples~\cite{Knapitsch2017tnt}, and Mip-NeRF360~\cite{barron2022mip} dataset. Also, we provide additional attention visualization in Fig.~\ref{fig:att_sup}. 
Further details for each example are provided in the corresponding captions in figures.

\vspace{15mm}

\clearpage

\clearpage

\end{document}